%% file: main.tex
\documentclass[letterpaper, 10 pt, journal, twoside]{IEEEtran}
\input{header.tex}
\definecolor{Gray}{gray}{0.85}
\newcolumntype{a}{>{\columncolor{Gray}}c}
\newcolumntype{b}{>{\columncolor{white}}c}
\makeatletter
\IEEEtriggercmd{\reset@font\normalfont\fontsize{5.0pt}{5.0pt}\selectfont}
\makeatother
\IEEEtriggeratref{1}

\author{Ruiqi Ni$^{1}$, Zherong Pan$^{2}$, and Xifeng Gao$^{2}$\vspace{-10px}%
\thanks{\footnotesize{
$^1$ Department of Computer Science, Florida State University, rn19g@my.fsu.edu. 
$^2$ Lightspeed \& Quantum Studios, Tencent America, $\{$zherong.pan.usa, gxf.xisha$\}$@gmail.com.
}}
}

\makeatletter
\newcommand\fs@ruled@notop{\def\@fs@cfont{\bfseries}\let\@fs@capt\floatc@ruled
  \def\@fs@pre{}%
  \def\@fs@post{\kern2pt\hrule\relax}%
  \def\@fs@mid{\kern2pt\hrule\kern2pt}%
  \let\@fs@iftopcapt\iftrue}
\renewcommand\fst@algorithm{\fs@ruled@notop}
\makeatother

\newif\ifarxiv
\arxivtrue

\title{Robust Multi-Robot Trajectory \revised{Optimization}
Using Alternating Direction Method of Multiplier\vspace{-10px}}
        
\allowdisplaybreaks
\begin{document}
\maketitle

\begin{abstract}
We propose a variant of alternating direction method of multiplier (ADMM) to solve constrained trajectory optimization problems. Our ADMM framework breaks a joint optimization into small sub-problems, leading to a low iteration cost and decentralized parameter updates. \revised{Starting from a collision-free initial trajectory, our method inherits the theoretical properties of primal interior point method (P-IPM), i.e., guaranteed collision avoidance and homotopy preservation throughout optimization,} while being orders of magnitude faster. We have analyzed the convergence and evaluated our method for time-optimal multi-UAV trajectory optimizations and simultaneous goal-reaching of multiple robot arms, where we take into consider kinematics-, dynamics-limits, and homotopy-preserving collision constraints. Our method highlights an order of magnitude's speedup, while generating trajectories of comparable qualities as state-of-the-art P-IPM solver.
\end{abstract}
\input{introduction.tex}
\input{related.tex}
\input{problem.tex}
\input{method.tex}

\input{evaluation.tex}
\input{conclusion.tex}
\AtNextBibliography{\scriptsize}
\printbibliography
\newpage
\ifarxiv\input{convergenceAM.tex}\else\fi
\ifarxiv\input{convergenceADMM1.tex}\else\fi
\end{document}

%% file: header.tex
\usepackage{amsmath,amsfonts,amssymb}
\let\labelindent\relax
\usepackage{prettyref}
\usepackage{mathrsfs}
\usepackage{graphicx}
\usepackage{wrapfig}
\usepackage{xcolor,colortbl}
\usepackage{subfig}
\usepackage{MnSymbol}
\usepackage{multirow}
\usepackage{booktabs}
\usepackage{algorithm}
\usepackage[noend]{algpseudocode}
\usepackage{enumitem}
\usepackage{stackengine}
\usepackage{siunitx}

\usepackage
[backend=bibtex,
bibstyle=ieee,
citestyle=numeric,
mincitenames=1,
maxcitenames=2,
natbib=true,
doi=false,
isbn=false,
url=false,
eprint=false]{biblatex}
\usepackage[colorlinks,allcolors=gray,hypertexnames=true]{hyperref} 
\newcommand{\citewithauthor}[1]{\citeauthor{#1} \cite{#1}}

\newtheorem{theorem}{\TE{Theorem}}[section]
\newtheorem{lemma}[theorem]{\TE{Lemma}}
\newtheorem{remark}[theorem]{\TE{Remark}}

\newtheorem{definition}[theorem]{\TE{Definition}}

\algnewcommand{\LineComment}[1]{\State \(\triangleright\) #1}
\algdef{SE}[DOWHILE]{Do}{doWhile}{\algorithmicdo}[1]{\algorithmicwhile\ #1}
\newcommand*{\colorboxed}{}
\def\colorboxed#1#{%
  \colorboxedAux{#1}%
}
\newcommand*{\colorboxedAux}[3]{%
  \begingroup
    \colorlet{cb@saved}{.}%
    \color#1{#2}%
    \boxed{%
      \color{cb@saved}%
      #3%
    }%
  \endgroup
}

\usepackage{xcolor}

\newcommand\numberthis{\addtocounter{equation}{1}\tag{\theequation}}
\newcommand{\labelthis}[1]{\numberthis\label{#1}}
\newrefformat{fig}{Figure~\ref{#1}}
\newrefformat{par}{Section~\ref{#1}}
\newrefformat{appen}{Appendix~\ref{#1}}
\newrefformat{sec}{Section~\ref{#1}}
\newrefformat{sub}{Section~\ref{#1}}
\newrefformat{table}{Table~\ref{#1}}
\newrefformat{ass}{Assumption~\ref{#1}}
\newrefformat{alg}{Algorithm~\ref{#1}}
\newrefformat{def}{Definition~\ref{#1}}
\newrefformat{thm}{Theorem~\ref{#1}}
\newrefformat{lem}{Lemma~\ref{#1}}
\newrefformat{step}{Step~\ref{#1}}
\newrefformat{ln}{Line~\ref{#1}}
\newrefformat{eq}{Equation~\ref{#1}}
\newrefformat{pb}{Problem~\ref{#1}}
\newrefformat{it}{Item~\ref{#1}}
\newrefformat{te}{Term~\ref{#1}}
\def\Eqref Eq:#1:{\eqref{eq:#1}}
\newrefformat{Eq}{Equation~\Eqref#1:}

\newcommand{\TE}[1]{\textbf{#1}}

\newcommand{\FPP}[2]{\frac{\partial{#1}}{\partial{#2}}}

\newcommand{\TWO}[2]{\left(\setlength{\arraycolsep}{1pt}\begin{array}{cc}{#1}, & {#2}\end{array}\right)}
\newcommand{\TWOC}[2]{\left(\setlength{\arraycolsep}{1pt}\begin{array}{c}#1 \\ #2\end{array}\right)}

\newcommand{\THREE}[3]{\left(\setlength{\arraycolsep}{1pt}\begin{array}{ccc}{#1}, & {#2}, & {#3}\end{array}\right)}

\newcommand{\THREEC}[3]{\left(\setlength{\arraycolsep}{1pt}\begin{array}{c}#1 \\ #2 \\ #3\end{array}\right)}

\newcommand{\FOUR}[4]{\left(\setlength{\arraycolsep}{1pt}\begin{array}{cccc}{#1}, & {#2}, & {#3}, & {#4}\end{array}\right)}

\newcommand{\FIVE}[5]{\left(\setlength{\arraycolsep}{1pt}\begin{array}{ccccc}{#1}, & {#2}, & {#3}, & {#4}, & {#5}\end{array}\right)}

\newcommand{\argmin}[1]{\underset{#1}{\text{argmin}}\;}

\newcommand{\ST}{\text{s.t.}}

\newcommand{\HULL}{\text{hull}}
\newcommand{\DIST}{\text{d}}
\newcommand{\FK}{\text{FK}}
\newcommand{\DOM}{\text{dom}}

\newcommand{\revised}[1]{\textcolor{black}{#1}}
\newcommand{\proofread}[1]{}
\newif\ifArxiv
\usepackage{xcolor}
\definecolor{Blue}{rgb}{1, 0.2, 0.8}

%% file: introduction.tex
\section{Introduction}
This paper focuses on trajectory optimization problems, a fundamental topic in robotic motion planning. Although the problem finds countless domains of applications, their pivotal common feature could be illustrated through the lens of two applications: goal-reaching of multiple UAVs and articulated robot arms. UAV trajectory optimization has been studied vastly \cite{goerzen2010survey}. Due to their small size and differential-flat dynamics \cite{lee2010geometric}, point-mass models can be used and Cartesian-space trajectories are linear functions of configuration variables. Furthermore, the quality of a UAV trajectory could be measured via convex metrics such as jerk or snap, casting trajectory optimization as convex programs. However, when flying in obstacle-rich environments and among other UAVs, non-convex, collision constraints must be considered \cite{augugliaro2012generation}. Failing to satisfy these constraints can render a generated trajectory completely useless. Handling articulated robot arms poses an even more challenging problem, where the linear dynamic assumption must be replaced with a nonlinear forward kinematic function that maps from configuration- to Cartesian-space, rendering all the Cartesian-space constraints non-convex. \revised{In summary, trajectory optimizer should pertain three properties: \TE{(versatility)} handle non-convex constraints and kinematic models; \TE{(robustness)} guarantee to satisfy all the constraints throughout optimization; \TE{(efficacy)} rapidly refine feasible initial trajectories into nearby, locally optimal solutions.}

\revised{We observe that prior trajectory optimization techniques exhibit remarkable performs under certain assumptions but have partial coverage of the three features above. For example, off-the-shelf primal-dual optimizers can solve general constrained programs and have been applied to trajectory optimization \cite{betts1993path,augugliaro2012generation,10.5555/3038546.3038571}. However, these methods violate robustness by allowing a feasible trajectory to leave the constraint manifold. Similarly, penalty methods \cite{doi:10.1177/0278364913488805,5980280,zhou2020ego2} have been used for trajectory optimization by replacing hard constraints with soft energies, which cannot guarantee robustness. On the other hand, we proposed a new optimizer in our prior work \cite{2010.09904} for UAV trajectory planning with perfect versatility and robustness, where all the constraints are converted into primal-only log-barrier functions with finite duality gap. As a result, all the constraints are satisfied throughout the optimization with the Continuous Collision Detection (CCD) bounded line search step. With the improved robustness, however, comes a significant sacrifice in efficacy. For the same benchmarks, our primal-only methods can take $3-5\times$ more computations to converge as compared with primal-dual counterparts. This is due to the log-barrier functions introducing arbitrarily large gradients near the constraint boundaries. As a result, an optimizer needs to use a costly line-search after each iteration to ensure a safe solution that satisfies all the stiff constraints. The gradient-flows of such objective functions are known as stiff dynamics, for which numerical time-integration can have ill-convergence as studied \cite{shampine1979user}.}

\TE{Main Results:} We propose a variant of ADMM-type solver that inherits the versatility and robustness from \cite{2010.09904}, while we achieve orders of magnitude higher performance. Intuitively, ADMM separates non-stiff and stiff objective terms into different sub-problems using slack variables, so that each sub-problem is well-conditioned. Moreover, since sub-problems are independent and involve very few decision variables, an ADMM iteration can be trivially parallelized and incurs a much lower cost. Existing convergence analysis, however, only guarantees that ADMM converges for convex problems or non-convex problems with \revised{linear or affine constraints \cite{gao2020admm}}. We present improved analysis which shows that our ADMM variant converges for both UAV and articulated trajectory planning problems under nonlinear collision constraints, kinematic- and dynamic-limits. We have applied our method to large-scale multi-UAV trajectory \revised{optimization} and articulated multi-robot goal-reaching problems as defined in \prettyref{sec:problem}. During our evaluations (\prettyref{sec:evaluation}), we observe tens of times' speedup over Newton-type methods. Our algorithms are detailed in \prettyref{sec:method}.

%% file: related.tex
\section{Related Work}
We review related trajectory generation techniques and cover necessary backgrounds in operations research.

\revised{\TE{Trajectory Generation} aims at computing robot trajectories from high-level goals and constraints.} Their typical scenarios of applications involve navigation \cite{panagou2014motion}, multi-UAV coordination \cite{choi2020multi}, human-robot interaction \cite{ragaglia2018trajectory}, tele-operation \cite{ardakani2014trajectory}, trajectory following \cite{5980409}, etc., where frequent trajectory update is a necessity to handle various sources of uncertainty. Due to the limited computational resources, early works use pre-computations to reduce the runtime cost. For example, \citewithauthor{panagou2014motion,shimoda2005potential} modulate a vector field to guide agents in a collision-free manner, while assuming point robots and known environments. Closed-form solutions such as \cite{qu2004new,shomin2013differentially} exist but are limited to certain types of dynamic systems or problem paradigms. More recently, \citewithauthor{belghith2006anytime,karaman2011anytime} have established anytime-variants of sampling-based roadmaps that continually improve an initial feasible solution during execution.

\revised{\TE{Trajectory Optimization} refines robot trajectories given a feasible or infeasible initial guess.} Trajectory optimization dates back to \cite{von1992direct,betts1993path}, but has recently gained significantly attention due to the maturity of nonlinear programming solvers. These methods have robots' goal of navigation formulated as objective functions, while taking various safety requirements as (non)linear constraints. They achieve unprecedented success in real-time control of high-dimensional articulated bodies \cite{todorov2012mujoco} and large swarms of UAVs \cite{augugliaro2012generation}. \revised{In particular, trajectory optimizer can also be used for trajectory generation by setting the initial trajectory to be a trivial solution \cite{todorov2012mujoco,9268217,zhou2020ego}.} On the down side, most trajectory optimization techniques suffer from a lack of robustness. Many works formulate constraints as soft objective functions \cite{8206214,todorov2012mujoco} or use primal-dual interior point methods to handle non-convex constraints \cite{betts1993path,8758904,zhou2020ego}, which does not guarantee constraint satisfaction. On the other hand, some techniques \cite{7839930,tordesillas2021faster} restrict the solution space to a disjoint convex subset so that efficient solvers are available, but these methods are limited to returning sub-optimal solutions. Instead, \revised{\citewithauthor{2010.09904} starts from a strictly feasible initial guess and uses a primal-only method to transform non-convex constraints into log-barrier functions with finite duality gap. Using a line-search with CCD as safe-guard, it is guaranteed to satisfy all constraints and no restrictions in solution space are needed.} But optimizer in this case can make slow progress, being blocked by the large gradient of log-barrier functions. This dilemma between efficacy and robustness has been studied as stiff dynamic systems \cite{shampine1979user}, for which dedicated techniques are developed for different applications such as numerical continuation \cite{vasudevan1989homotopy}. Unlike these methods, we show that first-order methods can be combined with barrier methods to achieve significant speedup.

\TE{Alternating Direction Method of Multiplier} is a first-order optimization framework originally designed in convex-programming paradigm \cite{boyd2011distributed}. ADMM features a low iteration cost and moderate accuracy of solutions, making it a stellar fit for trajectory optimization where many iterations can be performed within a short period of time. ADMM uses slack variables to split the problem into small subproblems and approximately maintains the consistency between slack and original variables by updating the Lagrangian multipliers. Although theoretical convergence guarantee was only available under convex assumptions, ADMM has been adapted to solve non-linear fluid dynamics control \cite{10.1145/3016963}, collision-free UAV trajectory generation \cite{9268217}, and bipedal locomotion \cite{zhou2020accelerated}, where good empirically performances have been observed. It was not until very recently that the good performances of ADMM in nonlinear settings have been theoretically explained by \cite{jiang2019structured,wang2019global,gao2020admm}. Unfortunately, even these latest analysis cannot cover many robotic applications such as \cite{9268217}. Our new ADMM algorithm correctly divides the responsibility between the slack and original variables, where the slack variables handle non-stiff function terms and the original variables handle stiff ones. By using line-search to ensure strict function decrease, we can prove convergence (speed) of our algorithm to a robust solution.

%% file: problem.tex
\section{Trajectory Optimization Paradigm\label{sec:problem}}
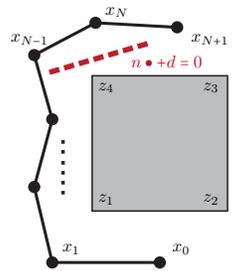
\begin{wrapfigure}{r}{0.15\textwidth}
\centering
\vspace{-5px}
\scalebox{0.65}{
\includegraphics[width=0.24\textwidth]{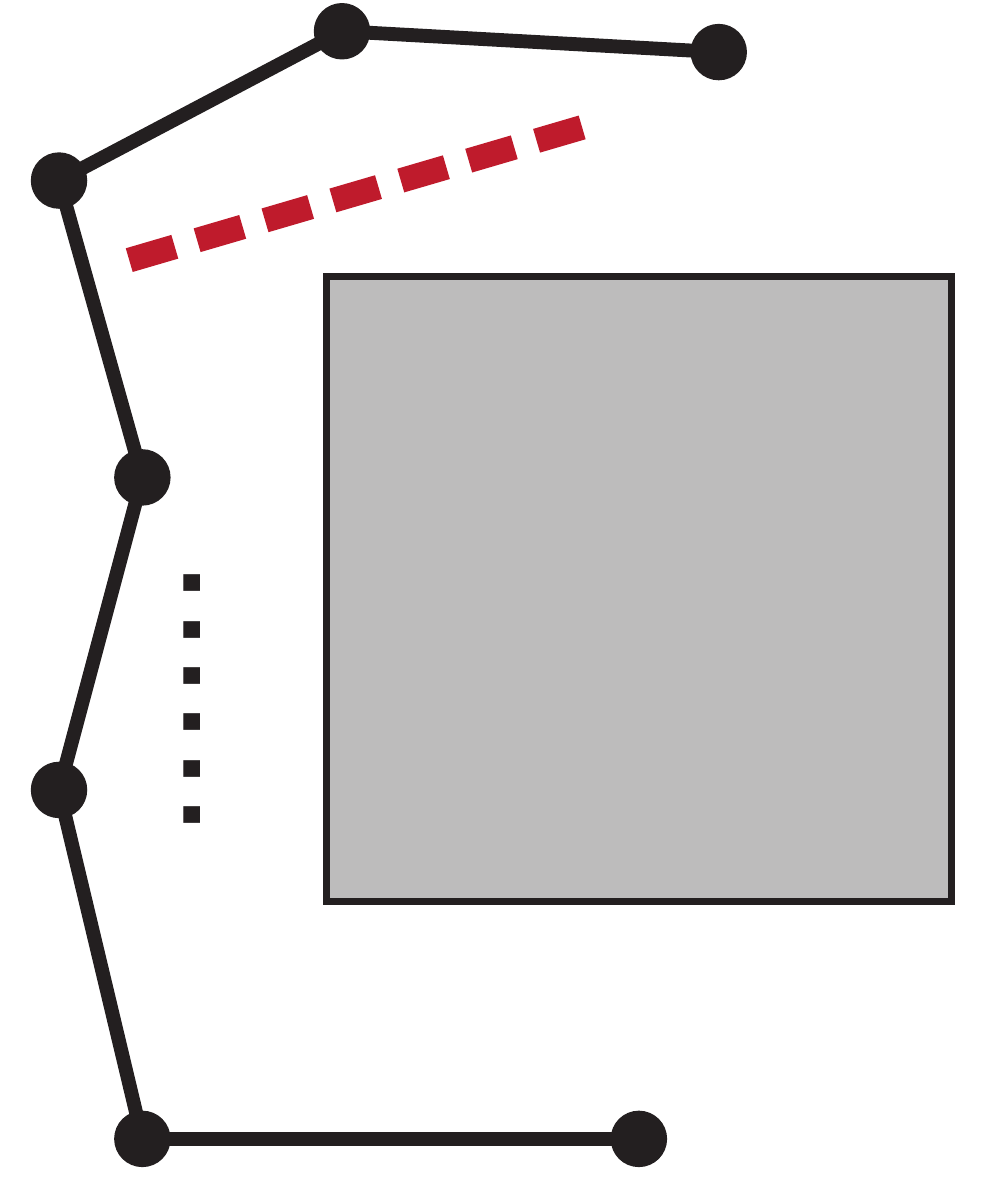}
\put(-79,42){$z_1$}
\put(-18,42){$z_2$}
\put(-79,107){$z_4$}
\put(-18,107){$z_3$}
\put(-36,13){$x_0$}
\put(-100,13){$x_1$}
\put(-130,135){$x_{N-1}$}
\put(-75,151){$x_{N}$}
\put(-25,135){$x_{N+1}$}
\definecolor{darkred}{RGB}{195,13,34}
\put(-60,120){\textcolor{darkred}{$n\bullet+d=0$}}}
\caption{\label{fig:prob} \small{Illustrative problem with the red dashed line being the separating plane.}}
\vspace{-5px}
\end{wrapfigure}
We motivate our analysis using a 2D collision-free trajectory optimization problem as illustrated in \prettyref{fig:prob}, \revised{while our algorithm is applied to both 2D and 3D workspaces alike.} Consider a point robot traveling in a 2D workspace along a piecewise linear trajectory discretized using $N+1$ linear segments and $N+2$ vertices. We further assume the start and goal positions are fixed, leaving the intermediary $N$ points as decision variables. This trajectory can be parameterized by a vector $x\in\mathbb{R}^{2N}$ where $x_i\in\mathbb{R}^2$ is the $i$th vertex. For simplicity in this example, our goal is for the trajectory to be as smooth as possible. Further, the robot must be collision-free and cannot intersect the box-shaped obstacle in the middle and we assume that the four vertices of the box are $Z=\{z_1,z_2,z_3,z_4\}$. These collision constraints can be expressed as:
\begin{align*}
\DIST(\HULL(X_i),\HULL(Z))\geq0\quad\forall i=1,\cdots,N-1,
\end{align*}
where $X_i=\{x_i,x_{i+1}\}$ is the $i$th line segment \revised{($X_i\in\mathbb{R}^4$ in this illustrative toy example)}, $\HULL(\bullet)$ denotes the convex hull and $\DIST$ is Euclidean the distance between two convex objects. A collision-free trajectory optimization problem can be formulated as:
\small
\begin{align}
\label{eq:prob1}
\argmin{x}\;&\sum_iX_i^TLX_i   \\
\ST\;&\DIST(\HULL(X_i),\HULL(Z))\geq0\quad\forall i=1,\cdots,N-1,\nonumber
\end{align}
\normalsize
where $L$ is the Laplacian stencil measuring smoothness and $X_i^TLX_i$ measures the squared length of $i$th line segment. However, \prettyref{eq:prob1} only considers geometric or kinematic constraints and a robot might not be able to traverse the optimized trajectory due to the violation physical constraints. In many problems, including autonomous driving \cite{katrakazas2015real} and UAV path planning \cite{cheng2008time}, a simplified physical model can be incorporated that only considers velocity and acceleration limits. We could approximate the velocity and acceleration using finite-difference as:
\begin{align*}
V_i\triangleq x_{i+1}-x_i\quad
A_i\triangleq x_{i+2}-2x_{i+1}+x_i,
\end{align*}
and formulate the time-optimal, collision-free trajectory optimization problem as:
\small
\begin{align}
\label{eq:prob2}
\argmin{x,\Delta t}\;&\sum_iX_i^TLX_i/\Delta t^2+w\Delta t   \\
\ST\;&\DIST(\HULL(X_i),\HULL(Z))\geq0\quad\forall i=1,\cdots,N-1\nonumber \\
&\|V_i\|\leq v_\text{max}\Delta t\quad\forall i=1,\cdots,N-1\nonumber   \\
&\|A_i\|\leq a_\text{max}\Delta t^2\quad\forall i=1,\cdots,N-2,\nonumber
\end{align}
\normalsize
where we use $\sum_iX_i^TLX_i/\Delta t^2$ to measure trajectory smoothness with time, e.g. Dirichlet energy, and we use a coefficient $w$ to balance between optimality in terms of trajectory length and arrival time. Here, ${v,a}_\text{max}$ are the upper bounds of velocities and accelerations. Although the above example is only considering a single robot and piecewise linear trajectories, extensions to several practical problem settings are straightforward, as discussed below.

\begin{figure}[ht]
\vspace{-5px}
\centering
\scalebox{0.55}{
\includegraphics[width=0.45\textwidth]{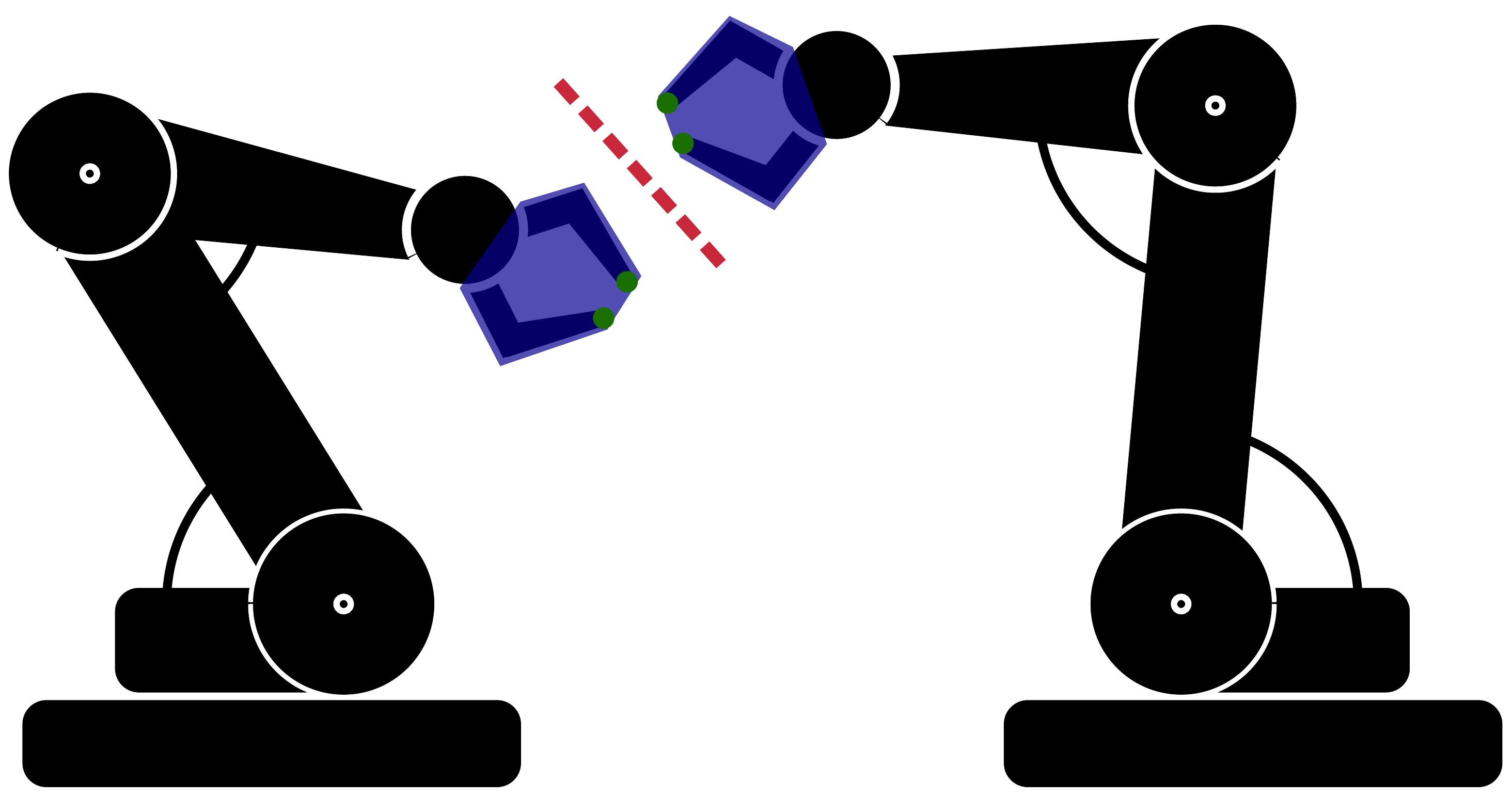}
\put(-215,40){$\theta_i$}
\put(-193,75){$\theta_i$}
\put(-22,40){$\theta_j$}
\put(-75,80){$\theta_j$}
\put(-138,65){$x_i$}
\put(-140,115){$x_j$}
\definecolor{darkred}{RGB}{195,13,34}
\put(-195,105){\textcolor{darkred}{$nx+d=0$}}}
\vspace{-5px}
\caption{\label{fig:ArmIllus}\small{\revised{The configuration of two robot arms includes joint angles $\theta_i,\theta_j$. The set of Cartesian points on the robot, $x_i,x_j$ (green), are mapped from $\theta_i,\theta_j$ using forward kinematic functions. To avoid collisions between the two robot arms, a convex hull is computed for each robot link (blue), and a separating plane (red) is used to ensure the pair of convex hulls stay on different sides.}}}
\vspace{-10px}
\end{figure}
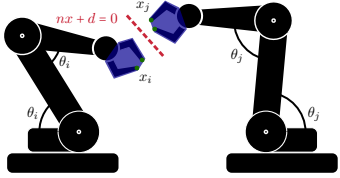
\subsection{Time-Optimal Multi-UAV \revised{Trajectory Optimization}}
We first extend our formulation to handle multiple UAV trajectories represented using composite B\'ezier curves with $N$ pieces of order $M$. In this case, the decision variable $x$ is a set of $N(M-2)+3$ control points or $x\in\mathbb{R}^{3(N(M-2)+3)}$ (note that the two neighboring curves share $3$ control points to ensure second order continuity and we refer readers to \cite{2010.09904} for more details). For the $i$th piece of B\'ezier curve, the velocity and acceleration are defined as:

\small
\begin{align*}
X_i\triangleq \THREEC{x_{i(M-2)-M+3}}{\vdots}{x_{i(M-2)+3}}\quad
V_i(s)\triangleq \dot{I}(s)X_i\quad
A_i(s)\triangleq \ddot{I}(s)X_i,
\end{align*}
\normalsize
where $X_i$ are the $M+1$ control points of the $i$th B\'ezier curve piece, and $I(s)$, $\dot{I}(s)$ and $\ddot{I}(s)$ are the B\'ezier curve's interpolation stencil for position, velocity and acceleration, respectively, with $s\in[0,1]$ being the natural parameter. It can be shown that $V_i(s),A_i(s)$ are also B\'ezier curves of orders $M-1$ and $M-2$, respectively. The velocity and acceleration limits must hold for every $s\in[0,1]$, for which a finite-dimensional, conservative approximation is to require all the control points of $\dot{I}(s)$ and $\ddot{I}(s)$ are bounded by $v_\text{max}$ and $a_\text{max}$. With a slight abuse of notation, we reuse $I,\dot{I},\ddot{I}$ without parameter $s$ to denote the matrices extracting the control points of $I(s)X_i,\dot{I}(s)X_i,\ddot{I}(s)X_i$, respectively. \revised{In other words, we define the vectors $V_i\triangleq\dot{I}X_i$ and $A_i\triangleq\ddot{I}X_i$ as the control points of the $i$th B\'ezier curve}, then the form of velocity and acceleration limits are identical to \prettyref{eq:prob2}. 

For multiple UAVs, however, we need to consider the additional collision constraints between different trajectories. To further unify the notations, we concatenate the control points of different UAVs into a single vector $x$, i.e., two B\'ezier curve pieces might correspond to different UAVs, and we introduce collision constraints between different UAVs:
\begin{align*}
\DIST(\HULL(X_i),\HULL(X_j))\geq0\quad\forall i,j\in\text{different UAV}.
\end{align*}
Our final formulation of time-optimal multi-UAV trajectory optimization takes the following form:
\small
\begin{align}
\label{eq:prob3}
\argmin{x,\Delta t}\;&\sum_i\mathcal{O}(X_i, \Delta t)  \\
\ST\;&\DIST(\HULL(X_i),\HULL(Z))\geq0\quad\forall i=1,\cdots,N-1\nonumber \\
&\DIST(\HULL(X_i),\HULL(X_j))\geq0\quad\forall i,j\in\text{different UAV}\nonumber  \\
&\|V_i\|\leq v_\text{max}\Delta t\quad\forall i=1,\cdots,N-1\nonumber\\
&\|A_i\|\leq a_\text{max}\Delta t^2\quad\forall i=1,\cdots,N-2,\nonumber
\end{align}
\normalsize
where we generalize the objective function with smoothness and time optimality to take an arbitrary, possibly non-convex form, $\mathcal{O}(X_i, \Delta t)$, which is a function of a single piece of sub-trajectory $X_i$. Almost all the objective functions in trajectory optimization applications can be written in this form. For example, smoothness can be written as the sum of total curvature, snap, or jerk of each piece, and the end-point cost is only related to the last piece.

\subsection{\revised{Goal-Reaching of} Articulated Robot Arms}
The position of UAV at any instance $s$ on the $i$th B\'ezier curve is \revised{$I(s)X_i$}, which is a linear function of decision variable $X_i$ and thus $x$. However, more general problem settings require non-linear relationships, of which a typical case is articulated robot arms as illustrated in \prettyref{fig:ArmIllus}. Consider the problem of multiple interacting robot arms in a shared 3D workspace. Each arm's Cartesian-space configuration at the $i$th time instance is represented by a triangle mesh with a set of vertices concatenated into the vector $x_i$. However, we need to maintain the corresponding configuration $\theta_i$, where $|\theta_i|$ is the degrees of freedom of each arm (DOF). Our decision variable is $\theta\in\mathbb{R}^{\text{DOF}\times N}$, each $x_i$ is a derived variable of $\theta_i$ via the forward kinematics function $x_i\triangleq\FK(\theta_i)$, and a linear interpolated Cartesian-space trajectory is: $X_i(\theta)=\left\{x_i,x_{i+1}\right\}=\left\{FK(\theta_i),FK(\theta_{i+1})\right\}$. We further define the velocity and acceleration in configuration space as:
\begin{align*}
V_i(\theta)\triangleq \theta_{i+1}-\theta_i\quad
A_i(\theta)\triangleq \theta_{i+2}-2\theta_{i+1}+\theta_i.
\end{align*}
In summary, the multi-arm goal-reaching problem can be formulated as:
\small
\begin{align}
\label{eq:prob4}
\argmin{\theta,\Delta t}\;&\sum_i\mathcal{O}(X_i(\theta), \Delta t)   \\
\ST\;&\DIST(\HULL(X_i(\theta)),\HULL(Z))\geq0\quad\forall i=1,\cdots,N-1\nonumber \\
&\DIST(\HULL(X_i(\theta)),\HULL(X_j(\theta)))\geq0\quad\forall i,j\in\text{different arm}\nonumber \\
&\|V_i(\theta)\|\leq v_\text{max}\Delta t\quad\forall i=1,\cdots,N-1\nonumber\\
&\|A_i(\theta)\|\leq a_\text{max}\Delta t^2\quad\forall i=1,\cdots,N-2.\nonumber
\end{align}
\normalsize
\prettyref{eq:prob4} takes a more general form than all the previous problems, and we would propose our variant of ADMM algorithm assuming this formulation. If our method is applied to UAV trajectory optimization, we can plug in the degenerate relationship $\FK(\theta_i)=\theta_i$.
\begin{remark}
\revised{By using separating planes to formulate collision constraints, we assume each robot link is convex. This treatment allows us to use only one separating plane between a pair of robot links. If concave features of robot links must be modeled accurately, then robot link must be further decomposed into convex parts, and a separating plane must be used between each pair of parts, leading to more separating planes and iterations.}
\end{remark}

%% file: method.tex
\section{ADMM-Type Trajectory Optimization\label{sec:method}}
ADMM is a variant of the Augmented Lagrangian Method (ALM) that does not update the penalty parameter. The main advantage of ADMM is that, by introducing slack variables, each substep consists of either a small problem involving the non-stiff part of the objective function or a large problem involving the stiff part of the objective function or constraints. As a result, the ADMM solver allows larger timestep sizes to be taken for the non-stiff part, leading to faster convergence. Although ADMM has only first-order convergence rate, it can quickly approximate a locally optimal solution with moderate accuracy, which is sufficiently for trajectory optimization. 

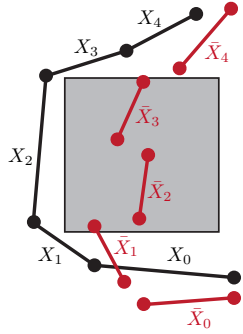
\begin{wrapfigure}{r}{0.18\textwidth}
\centering
\vspace{-10px}
\scalebox{0.75}{
\includegraphics[width=0.24\textwidth]{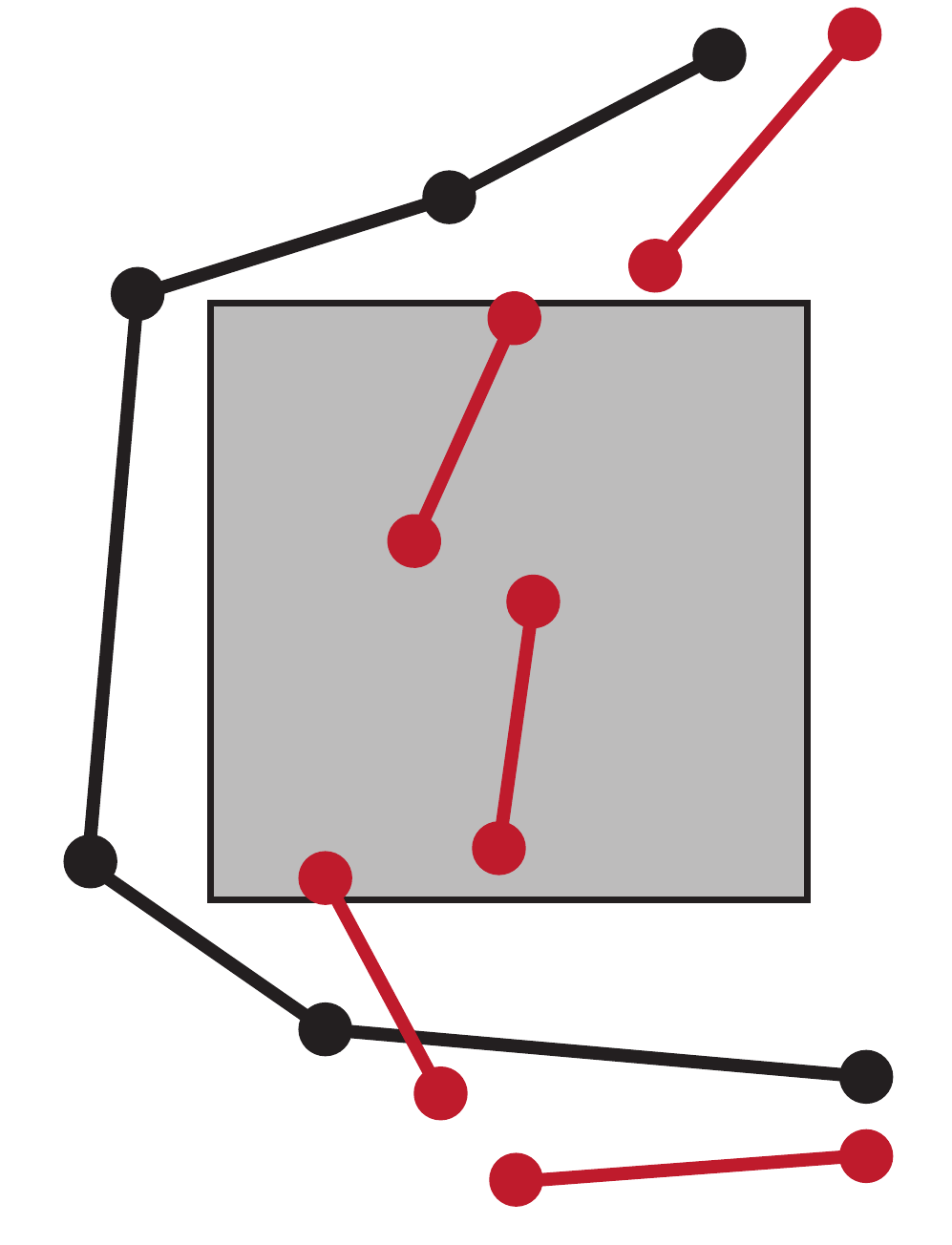}
\put(-45,30){$X_0$}
\put(-110,30){$X_1$}
\put(-125,80){$X_2$}
\put(-92,135){$X_3$}
\put(-60,150){$X_4$}
\definecolor{darkred}{RGB}{195,13,34}
\put(-35,-2){\textcolor{darkred}{$\bar{X}_0$}}
\put(-72,34){\textcolor{darkred}{$\bar{X}_1$}}
\put(-55,62){\textcolor{darkred}{$\bar{X}_2$}}
\put(-61,100){\textcolor{darkred}{$\bar{X}_3$}}
\put(-26,132){\textcolor{darkred}{$\bar{X}_4$}}}
\caption{\label{fig:consistency} \small{ADMM maintain two sets of variables $X_i$ and $\bar{X}_i$. \revised{$X_i$ satisfies all constraints throughout optimization, while $\bar{X}_i$ can violate constraints during intermediary iterations. On convergence, however, both $X_i$ and $\bar{X}_i$ satisfy all constraints.}}}
\vspace{-5px}
\end{wrapfigure}
ADMM handles inequality constraints by reformulating them as indicator functions, but we are handling possibly non-convex constraints for which projection operators, which is associated with indicator functions, do not have closed-form solutions. Instead, we follow our prior work \cite{2010.09904} and rely on a log-barrier relaxations with non-zero duality gap. For example, if we have a hard constraint $g(x)\geq0$ where $g$ is some differentiable function, then the feasible domain can be identified with the finite sub-level set of the log-barrier function: $-\text{log}(g(x))$. We apply this technique to the velocity and acceleration limits. A similar technique can be used for collision constraints with the help of a separating plane, which is illustrated in \prettyref{fig:prob} as $n^T\bullet+d=0$ ($n$ is the plane normal, $d$ is the position, and $\bullet$ is the arbitrary point on the plane). Since we only consider distance between convex hulls, two convex hulls are non-overlapping if and only if there is a separating plane such that the two hulls are on different sides. We propose to optimize the parameters of the separating plane ($n,d$) as additional slack variables. As a result, the collision constraints become convex when fixing $n,d$ and optimizing the trajectory alone. Applying this idea to all the constraints and we can transform \prettyref{eq:prob4} into the following unconstrained optimization:

\small
\begin{align*}
&\argmin{\substack{\theta,\Delta t,d_i,d_{ij}\\\|n_i\|,\|n_{ij}\|=1}}\;\mathcal{L}(\theta,\Delta t,n_i,d_i,n_{ij},d_{ij})\triangleq\sum_i\mathcal{O}(X_i(\theta),\Delta t)-  \\
&\gamma\sum_{i}\log(v_\text{max}\Delta t-\|V_i\|)-\gamma\sum_{i}\log(a_\text{max}\Delta t^2-\|A_i\|)-  \\
&\gamma\sum_{i}\left[\sum_{x\in X_i}\log(n_ix(\theta)+d_i)+\sum_{z\in Z}\log(-n_iz-d_i)\right]-    \\
&\gamma\sum_{ij}\left[\sum_{x\in X_i}\log(n_{ij}x(\theta)+d_{ij})+\sum_{x\in X_j}\log(-n_{ij}x(\theta)-d_{ij})\right],\labelthis{eq:prob4Log}
\end{align*}
\normalsize
where $\gamma$ is the weight of log-barrier function that can be tuned for each problem to control the exactness of constraint satisfaction. \revised{We use the same subscript convention as \prettyref{sec:problem}. Specifically, for $i$th trajectory piece $X_i$ and the obstacle $Z$, we introduce a separating plane $n_i^T\bullet+d_i=0$. For the pair of $i$th and $j$th trajectories pieces that might collide, we introduce a separating plane $n_{ij}^T\bullet+d_{ij}=0$.}

\prettyref{eq:prob4Log} is a strongly coupled problem with six sets of decision variables, where the constraints (or log-barrier functions) and the objective $\mathcal{O}$ are added up. However, these two kinds of functions have very different properties. The log-barrier functions are ``stiff'' and do not have a Lipschitz constant, which could generate arbitrarily large blocking gradients near the constraint boundaries, but the objective $\mathcal{O}$ is well-conditioned, oftentimes having a finite Lipschitz constant. Our main idea is to handle these functions in separate subproblems.

\begin{algorithm}[ht]
\caption{\label{alg:AM} AM}
\small{\begin{algorithmic}[1]
\Require{$\theta^0,\Delta t^0,\TWO{n_i}{d_i}^0,\TWO{n_{ij}}{d_{ij}}^0$}
\For{$k=0,1,\cdots$}\Comment{Update $\theta,\Delta t$}
\State $\theta^{k+1},\Delta t^{k+1}\approx\argmin{\theta,\Delta t}\mathcal{L}(\theta,\Delta t,n_i^k,d_i^k,n_{ij}^k,d_{ij}^k)$
\label{ln:AMProb1}
\For{collision constraint between $i$th piece of trajectory and environment}\Comment{Optimize separating plane}
\State $\TWO{n_i}{d_i}^{k+1}\approx\argmin{\|n_i\|=1,d_i}\mathcal{L}(\theta^{k+1},\Delta t^{k+1},n_i,d_i,n_{ij}^k,d_{ij}^k)$
\label{ln:AMProb2}
\EndFor
\For{collision constraint between $i$th and $j$th piece of trajectory}
\State $\TWO{n_{ij}}{d_{ij}}^{k+1}\approx\argmin{\|n_{ij}\|=1,d_{ij}}$
\State $\mathcal{L}(\theta^{k+1},\Delta t^{k+1},n_i^{k+1},d_i^{k+1},n_{ij},d_{ij})$
\label{ln:AMProb3}
\EndFor
\EndFor
\end{algorithmic}}
\end{algorithm}
\subsection{Alternating Minimization (AM)}
Before we describe our ADMM-type method, we review the basic alternating minimization scheme. AM has been used in trajectory optimization to handle time-optimality \cite{wang2020alternating} and collision constraints \cite{8424034}. A similar method can be applied to minimize \prettyref{eq:prob4Log} that alternates between updating the separating plane $n_i,d_i,n_{ij},d_{ij}$ and the robot configurations $\theta$ as outlined in AM \ref{alg:AM}. AM can be used along with ADMM while being easier to analyze. Prior work \cite{8424034} did not provide a convergence analysis and \citewithauthor{wang2020alternating} setup the first order convergence for a specific, strictly convex objective function where each minimization subproblem is a single-valued map. In \ifarxiv \prettyref{sec:convAM}, \else the appendices of our arxiv version \cite{2111.07016}, \fi we establish the convergence of AM \ref{alg:AM} for twice-differentiable objective functions with plane normals $n_i,n_{ij}$ constrained to the unit circle/sphere. In the next section, we will combine AM and ADMM, specifically we update robot configurations $\theta$ using ADMM and update separating planes $n_i,d_i,n_{ij},d_{ij}$ using AM.

\input{methodPartial.tex}

%% file: methodPartial.tex
\begin{algorithm}[ht]
\caption{\label{alg:ADMM1} ADMM with Stiffness Decoupling}
\small{\begin{algorithmic}[1]
\Require{$\theta^0,\Delta t^0,\Delta \bar{t}_i^0,\bar{X}_i^0,\lambda_i^0,\TWO{n_i}{d_i}^0,\TWO{n_{ij}}{d_{ij}}^0$}
\For{$k=0,1,\cdots$}\Comment{Update $\theta,\Delta t,\bar{X}_i,\lambda_i$}
\State $\theta^{k+1},\Delta t^{k+1}\approx\argmin{\theta,\Delta t}\mathcal{L}(\theta,\Delta t,\Delta \bar{t}_i,\bar{X}_i^k,\lambda_i^k,n_i^k,d_i^k,n_{ij}^k,d_{ij}^k)$
\For{$i$th piece of trajectory}
\State $\Delta \bar{t}_i^{k+1},\bar{X}_i^{k+1}\approx\argmin{\Delta \bar{t}_i,\bar{X}_i}$
\State $\mathcal{L}(\theta^{k+1},\Delta t^{k+1},\Delta \bar{t}_i,\bar{X}_i,\lambda_i^k,n_i^k,d_i^k,n_{ij}^k,d_{ij}^k)+$
\State $\frac{\varrho}{2}\|\bar{X}_i-\bar{X}_i^k\|^2$
\label{ln:ADMMProb1}
\State $\lambda_i^{k+1}\gets\lambda_i^k+\varrho(X_i(\theta^{k+1})-\bar{X}_i^{k+1})$
\label{ln:ADMMProb2}
\State $\Lambda_i^{k+1}\gets\Lambda_i^k+\varrho(\Delta t^{k+1}-\Delta \bar{t}_i^{k+1})$
\EndFor
\For{collision-free constraint between $i$th piece of trajectory and environment}\Comment{Optimize separating plane}
\State $\TWO{n_i}{d_i}^{k+1}\approx\argmin{\|n_i\|=1,d_i}$
\State $\mathcal{L}(\theta^{k+1},\Delta t^{k+1},\Delta \bar{t}_i^{k+1},\bar{X}_i^{k+1},\lambda_i^{k+1},n_i,d_i,n_{ij}^k,d_{ij}^k)$
\EndFor
\For{collision-free constraint between $i$th and $j$th piece of trajectory}
\State $\TWO{n_{ij}}{d_{ij}}^{k+1}\approx\argmin{\|n_{ij}\|=1,d_{ij}}$
\State $\mathcal{L}(\theta^{k+1},\Delta t^{k+1},\Delta \bar{t}_i^{k+1},\bar{X}_i^{k+1},\lambda_i^{k+1},n_i^{k+1},d_i^{k+1},n_{ij},d_{ij})$
\EndFor
\EndFor
\end{algorithmic}}
\end{algorithm}
\subsection{ADMM with Stiffness Decoupling}
The key idea behind ADMM is to treat stiff and non-stiff functions separately by introducing slack variables. Specifically, we introduce slack variables $\bar{X}_i$ for each $i=1,\cdots,N$ and transforms \prettyref{eq:prob4Log} into the following equivalent form:

\small
\begin{align*}
&\argmin{\substack{\theta,\Delta t,\Delta \bar{t}_i,\Bar{X}_i\\
d_i,d_{ij},\|n_i\|,\|n_{ij}\|=1}}\; \sum_i\left[\mathcal{O}(\bar{X}_i,\Delta \bar{t}_i)\right]-\labelthis{eq:prob4LogADMM}    \\
&\gamma\sum_{i}\log(v_\text{max}\Delta t-\|V_i\|)-\sum_{i}\gamma\log(a_\text{max}\Delta t^2-\|A_i\|)-    \\
&\gamma\sum_{i}\left[\sum_{x\in X_i}\log(n_ix(\theta)+d_i)+\sum_{z\in Z}\log(-n_iz-d_i)\right]-    \\
&\gamma\sum_{ij}\left[\sum_{x\in X_i}\log(n_{ij}x(\theta)+d_{ij})+\sum_{x\in X_j}\log(-n_{ij}x(\theta)-d_{ij})\right]   \\
&\ST\;X_i(\theta)=\bar{X}_i\land \Delta t=\Delta \bar{t}_i.
\end{align*}
\normalsize
By convention, we use a bar to indicate slack variables, i.e. $\bar{V}_i(s)=\dot{A}(s)\bar{X}_i$ and $\bar{A}_i(s)=\ddot{A}(s)\bar{X}_i$. 
\begin{remark}
We choose to have all slack variables reside in Cartesian space. As a result, if non-linear forward kinematics functions are used, ADMM must handle nonlinear constraint $X_i(\theta)=\bar{X}_i$.
\end{remark}
ADMM proceeds by transforming the equality constraints in \prettyref{eq:prob4LogADMM} into augmented Lagrangian terms. We arrive at the following augmented Lagrangian function:

\footnotesize
\begin{align*}
&\mathcal{L}(\theta,\Delta t,\Delta \bar{t}_i,\bar{X}_i,\lambda_i,n_i,d_i,n_{ij},d_{ij})\triangleq\sum_i\mathcal{O}(\bar{X}_i,\Delta \bar{t}_i)-    \\
&\gamma\sum_{i}\log(v_\text{max}\Delta t-\|V_i\|)-\gamma\sum_{i}\log(a_\text{max}\Delta t^2-\|A_i\|)-    \\
&\gamma\sum_{i}\left[\sum_{x\in X_i}\log(n_ix(\theta)+d_i)+\sum_{z\in Z}\log(-n_iz-d_i)\right]-    \\
&\gamma\sum_{ij}\left[\sum_{x\in X_i}\log(n_{ij}x(\theta)+d_{ij})+\sum_{x\in X_j}\log(-n_{ij}x(\theta)-d_{ij})\right]+   \\
&\sum_i\frac{\varrho}{2}\|X_i(\theta)-\bar{X}_i\|^2+\lambda_i^T(X_i(\theta)-\bar{X}_i)+\\
&\sum_i\frac{\varrho}{2}\|\Delta t-\Delta \bar{t}_i\|^2+\Lambda_i^T(\Delta t-\Delta \bar{t}_i),\labelthis{eq:ALF}
\end{align*}
\normalsize
where $\varrho$ is the penalty parameter, $\lambda_i$ is the augmented Lagrangian multiplier for $\bar{X}_i$. We can now present our ADMM algorithm seeking stationary points of \prettyref{eq:ALF}. Each iteration of our ADMM \ref{alg:ADMM1} is a five-way update that alternates between $\{\theta,\Delta t\},\{\Delta \bar{t}_i,\bar{X}_i\},\{\lambda_i,\Lambda_i\},\TWO{n_i}{d_i},\TWO{n_{ij}}{d_{ij}}$. Note that our objective function only appears in the $\{\Delta \bar{t}_i,\bar{X}_i\}$-subproblem, which does not involve any stiff, log-barrier functions. Therefore, ADMM \ref{alg:ADMM1} achieves stiffness decoupling.
\begin{remark}
For each optimization subproblem of ADMM \ref{alg:ADMM1}, we assume that the decision variable is initialized from last iteration. \revised{In \ifarxiv \prettyref{sec:convAM},\ref{sec:convADMM1}, \else the appendices of our arxiv version \cite{2111.07016}, \fi we show that subproblems in AM \ref{alg:AM} and ADMM \ref{alg:ADMM1} only need to be solved approximately. Specifically, we update $\theta,\Delta t,n_i,d_i,n_{ij},d_{ij}$ using a single (Riemannian) line search step, and we update $\Delta\bar{t},\bar{X}$ using a linearized function $\mathcal{L}$. For brevity, we denote these approximate oracles using an $\approx$ symbol.}
\end{remark}

\begin{figure*}
\centering
\scalebox{0.9}{
\includegraphics[width=.64\textwidth]{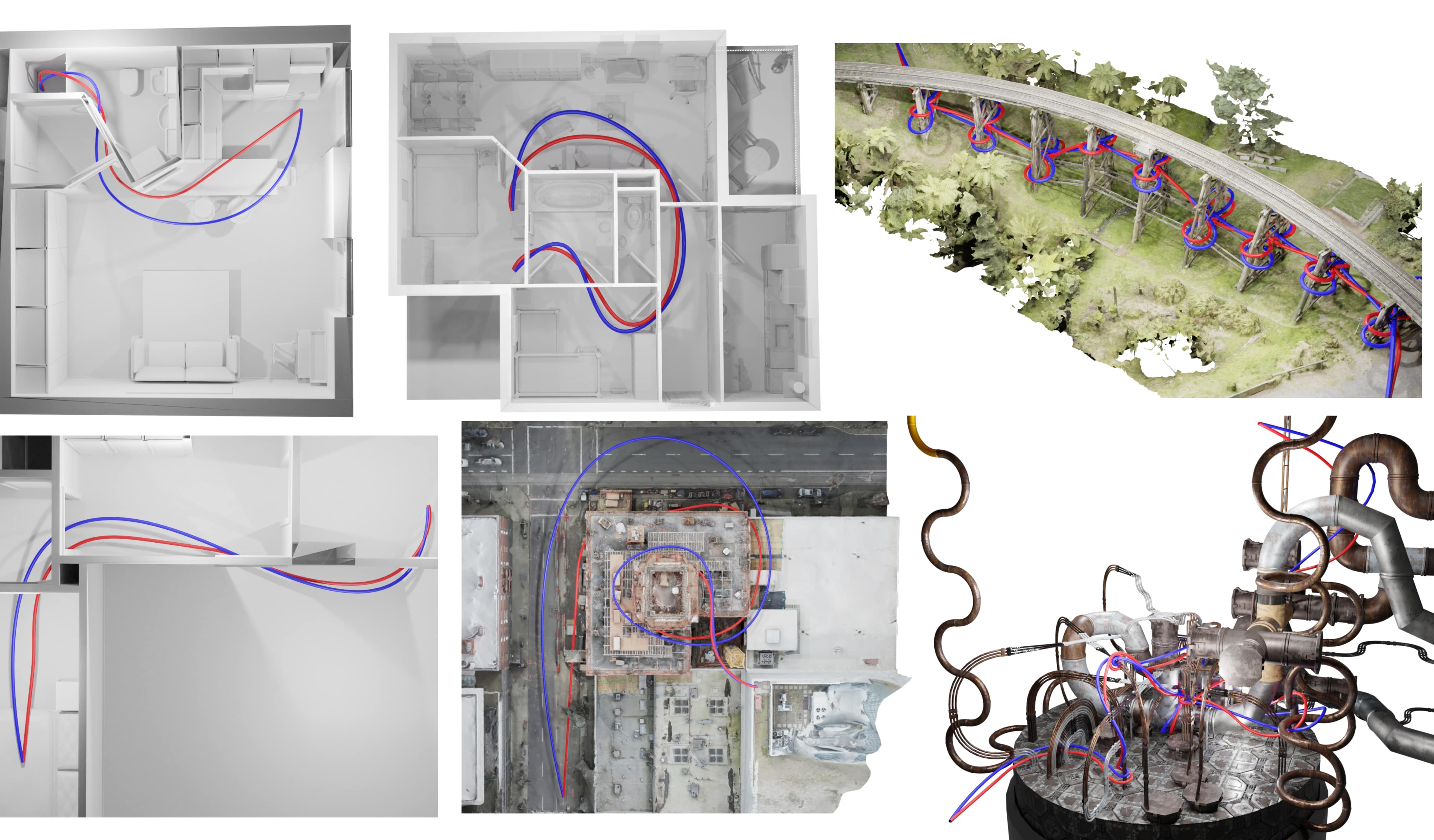}
\def\big{\includegraphics[width=.33\textwidth]{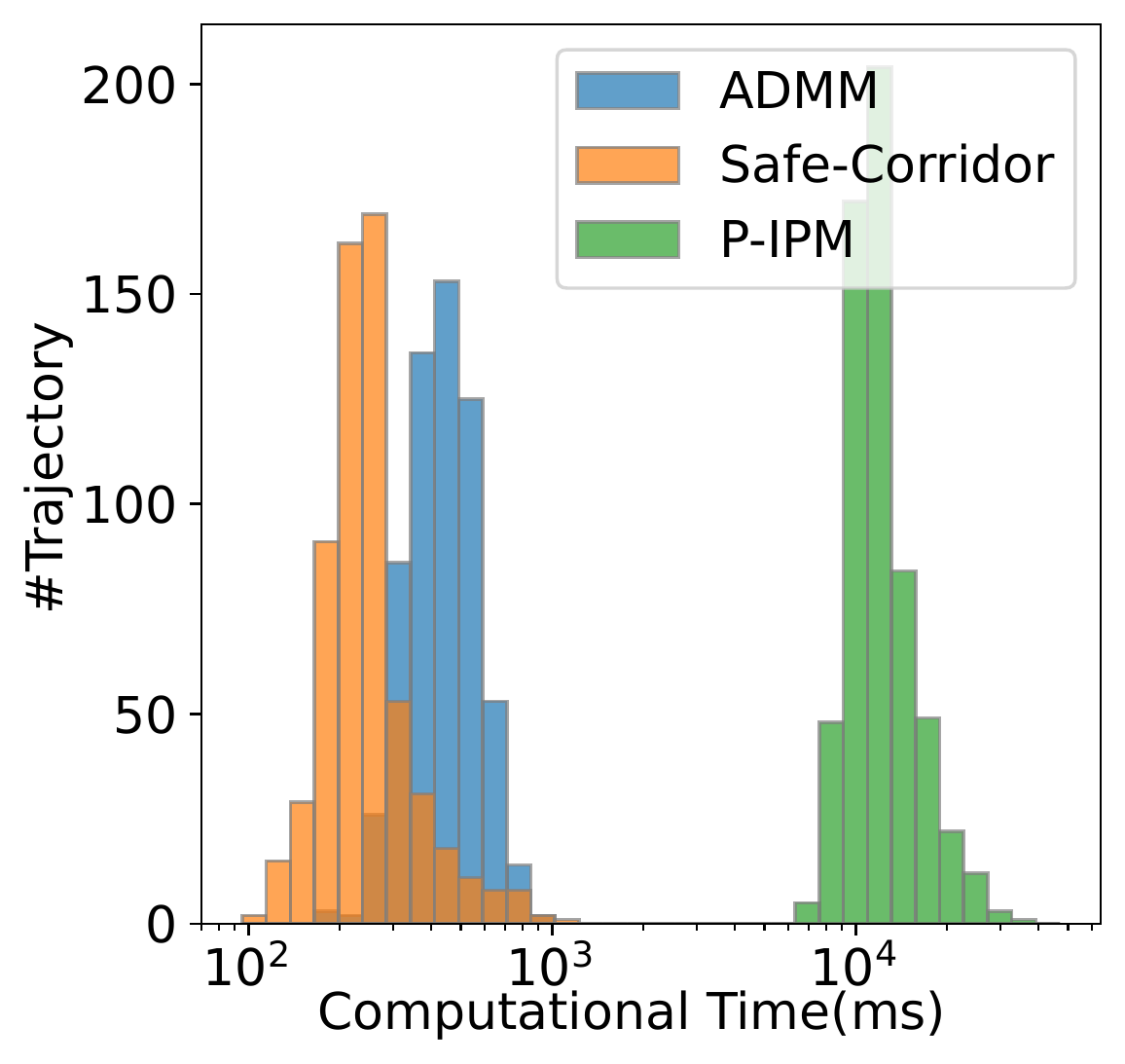}}
\def\little{\includegraphics[width=.09\textwidth,trim=3.5cm -6.5cm 3.5cm -6.5cm,clip]{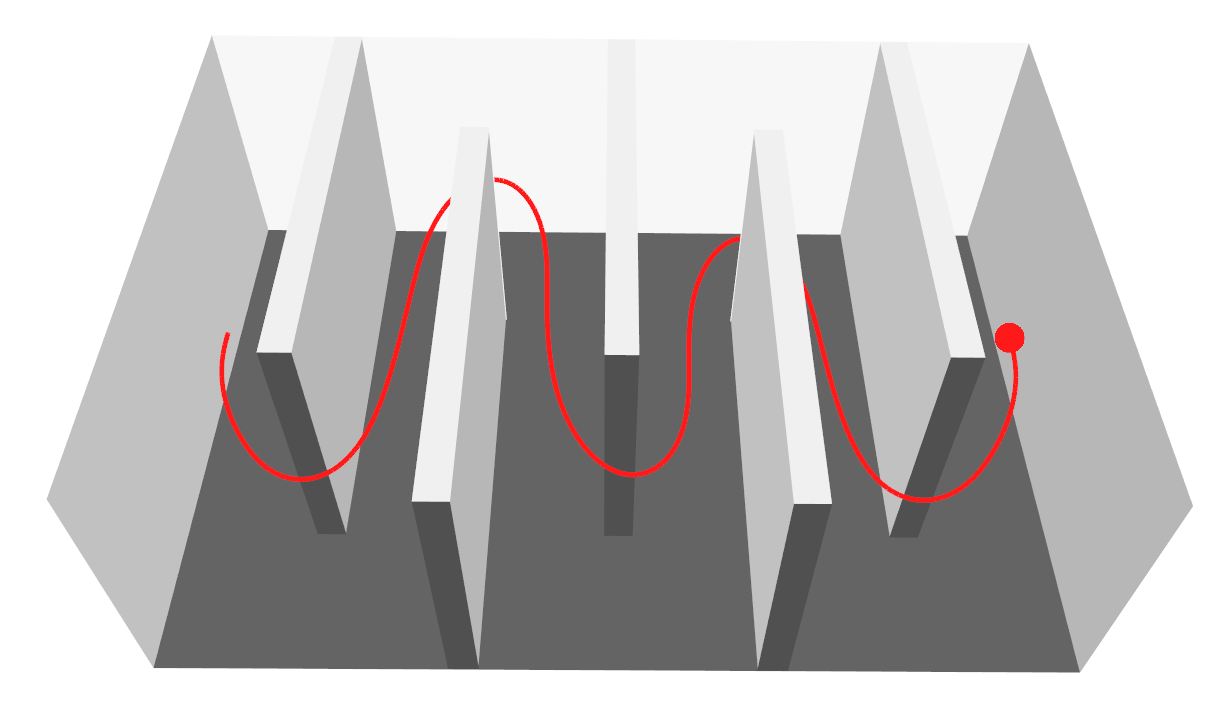}}
\stackinset{c}{15pt}{c}{10pt}{\little}{\big}
\put(-480,118){(a)}
\put(-345,118){(c)}
\put(-295,118){(f)}
\put(-480,10){(b)}
\put(-335,10){(d)}
\put(-275,60){(e)}
\put(-75 ,40){(g)}
\vspace{-10px}
}
\hfill
\captionof{figure}{\small{Examples of trajectories generated by Safe-Corridor \cite{han2021fast} (blue) and ADMM \ref{alg:ADMM1} (red) for a single UAV in complex environments: indoor flight (a-c), outdoor flight (a-f). \revised{(g): The distributions of computational time of ADMM \ref{alg:ADMM1}, P-IPM \cite{2010.09904}, and Safe-Corridor \cite{han2021fast} over $600$ trajectories computed for a synthetic environment in which we compute $600$ random initial trajectories using RRT-connect.}}}
\label{fig:singleUAV}
\vspace{-10px}
\end{figure*}
\begin{figure}[h]
\vspace{-5px}
\centering
\includegraphics[width=.48\textwidth]{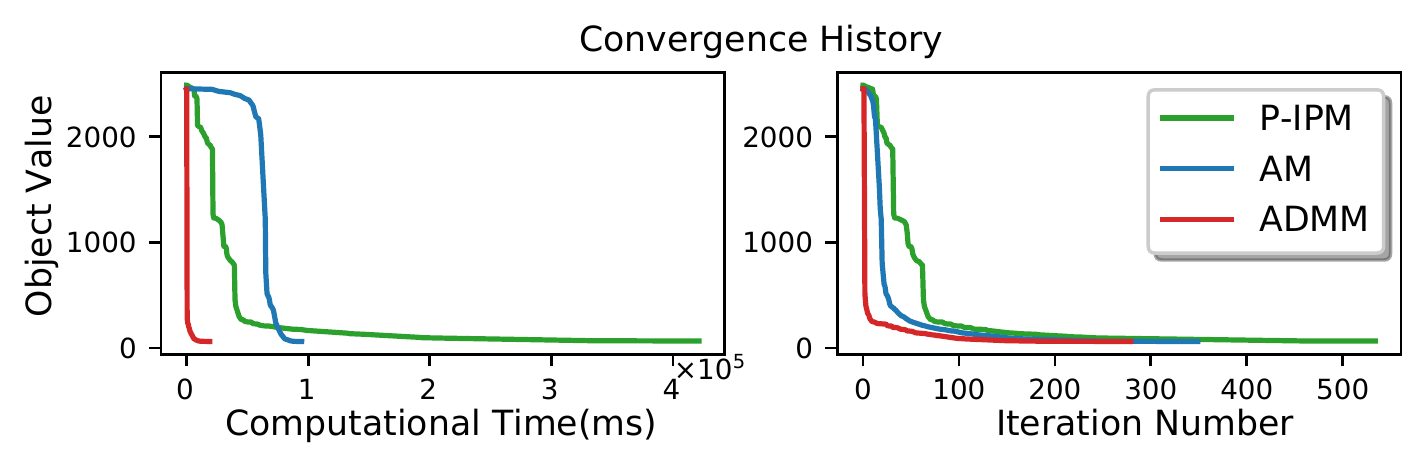}
\put(-150,26){(a)}
\put(-35,26){(b)}
\vspace{-5px}
\caption{\label{fig:UAVConv}\small{Convergence history of various algorithms for the example in \prettyref{fig:singleUAV} (f), comparing AM, ADMM, and Newton-type method (P-IPM) in terms of wall-time (a) and \#Iterations (b).}}
\vspace{-15px}
\end{figure}

\begin{remark}
ADMM always maintains two representations of the trajectory, $X_i$ and $\bar{X}_i$, where $X_i$ is used to satisfy the collision constraints and $\bar{X}_i$ (the slack variable) focuses on minimizing the objective function $\mathcal{O}$ at the risk of violating the collision constraints, as illustrated in \prettyref{fig:consistency}. \revised{It is know that using slack variables can loosen constraint satisfaction. As a result, we choose to formulate collision constraints and other hard constraints on $X_i$ instead of $\bar{X}_i$, so that all the constraints can be satisfied using a line-search step on $X_i$ variables. For collision constraints in particular, we inherit the line-step technique from \cite{2010.09904} that is safe-guarded by CCD. The CCD procedure ensures that there always exists a separating plane to split each pair of convex objects and the Lagrangian function $\mathcal{L}$ always takes a finite value.} On convergence, however, the two representations coincide and both collision-free and local optimal conditions hold.
\end{remark}

\begin{remark}
When updating the separating plane, the normal vector must be constrained to have unit length. These constraints can be reparameterized as an optimization on $\mathcal{SO}(3)$. Specifically, given the current solution denoted as $n_{ij}^{curr}$ with $\|n_{ij}^{curr}\|=1$, we reparameterize $n_{ij}$ by pre-multiplying a rotation matrix $R=\exp(r)$ by $n_{ij}^{curr}$, where we use the Rodriguez formula to parameterize a rotation matrix as the exponential of an arbitrary 3-dimensional vector $r$. Instead of using $n_{ij}$ as decision variables, we let $n_{ij}=\exp(r)n_{ij}^{curr}$ and use $r$ as our decision variables. Whichever value $r$ takes, we can ensure $\|n_{ij}\|=1$ (we refer reads to \cite{taylor1994minimization} for more details).
\end{remark}

%% file: evaluation.tex
\section{Evaluations\label{sec:evaluation}}
Our implementation uses C++11. Experiments are performed on a workstation with a 3.5 GHz Intel Core i9 processor. For experiments, we choose a unified set of parameters $v_{max}=2m/s, a_{max}=2m/s^2$ for UAVs (resp. $v_{max}=0.1m/s, a_{max}=0.1m/s^2$ for articulated bodies), $w=10^8, \varrho=0.1$ unless otherwise stated. \revised{We use the same weight, $\gamma=10$, for all the log-barrier functions. Although fine-tuning $\gamma$ separately for each log-barrier function can lead to better results, we find the same $\gamma=10$ achieves reasonably good results over all examples.} We use a locally supported log-barrier function as done in our prior work \cite{2010.09904}, which is active only when the distance between two objects is less than $0.1m$ for UAVs (resp. $0.04m$ for articulated bodies). Further, we set our clearance distance to be $0.1m$ for UAVs (resp. $10^{-3}m$ for articulated bodies), which can be plugged into CCD used by our line-search algorithm. Our algorithm terminates when $\|\nabla_{\theta^{k+1}}\mathcal{L}\|_\infty<\epsilon$ and we choose $\epsilon=10^{-2}$ for UAVs (resp. $\epsilon=10^{-1}$ for articulated bodies). By comparing with our prior work \cite{2010.09904}, we demonstrate the ability of ADMM in terms of resolving stiff-coupling issues and boosting the overall perform. We further compare with prior works \cite{zhou2020ego2,han2021fast} to highlight the robustness of our approach.

\begin{figure*}
\centering
\scalebox{0.9}{
\includegraphics[width=0.64\textwidth]{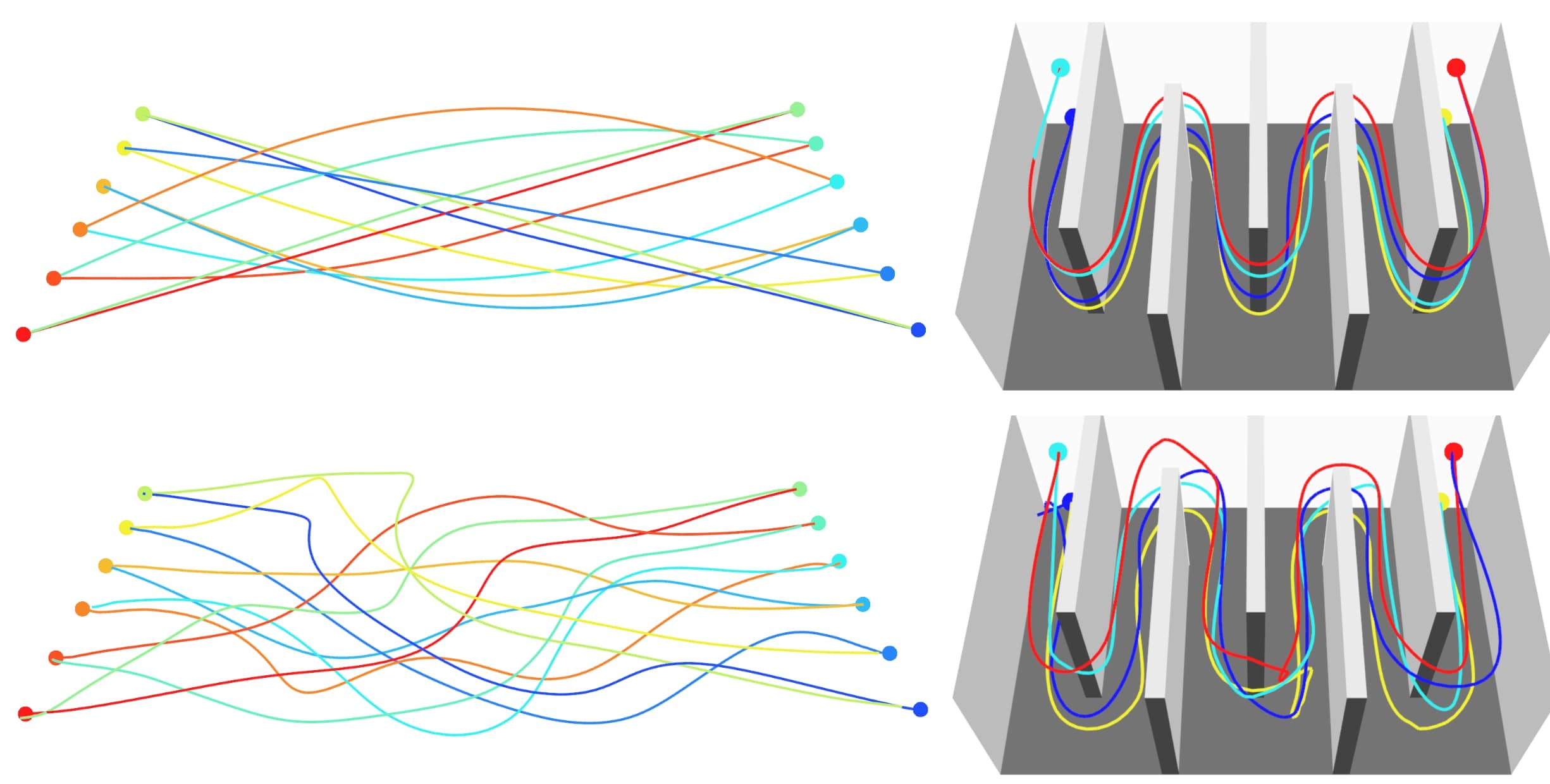}
\def\big{\includegraphics[width=.33\textwidth]{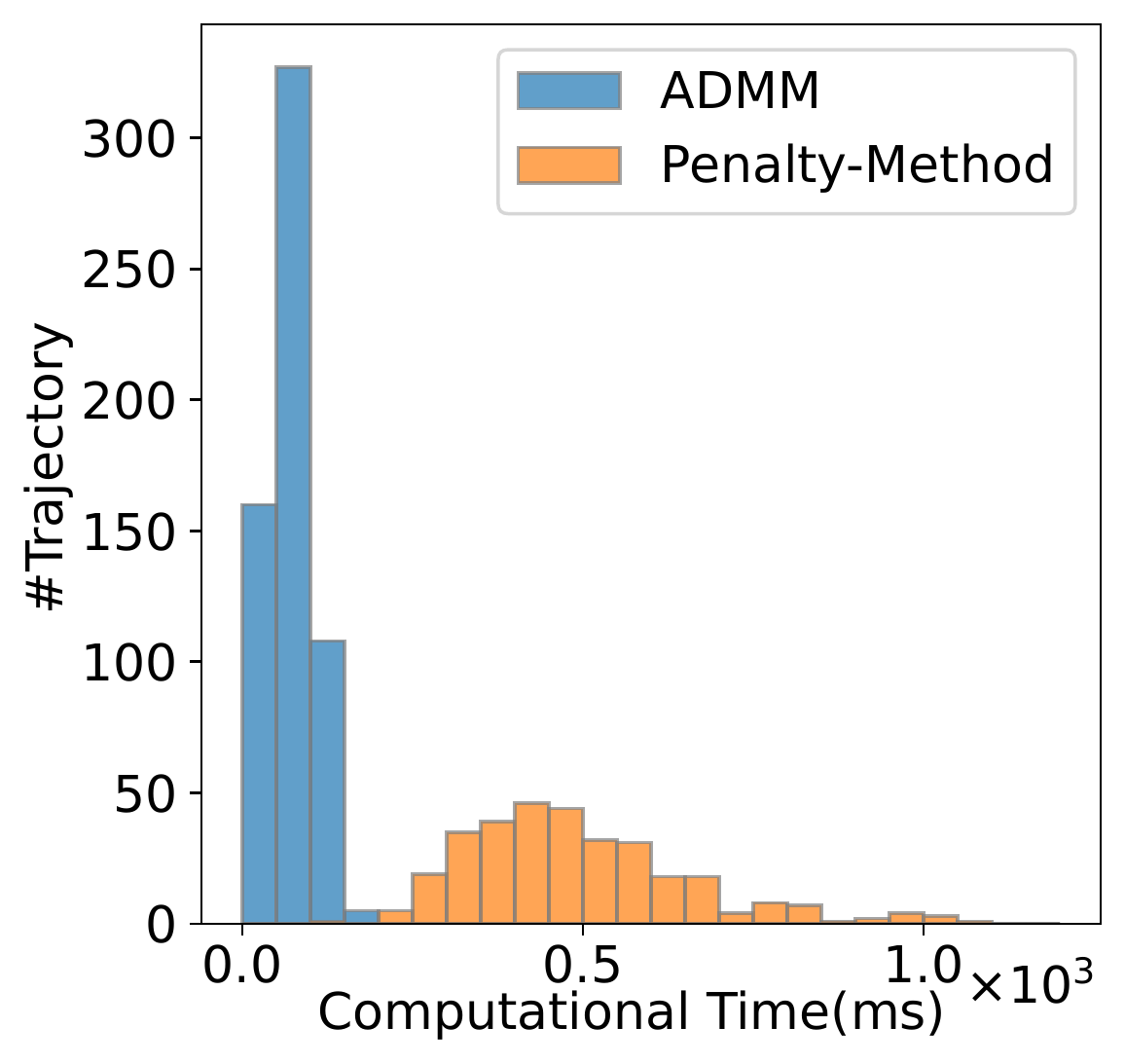}}
\def\little{\includegraphics[width=.12\textwidth]{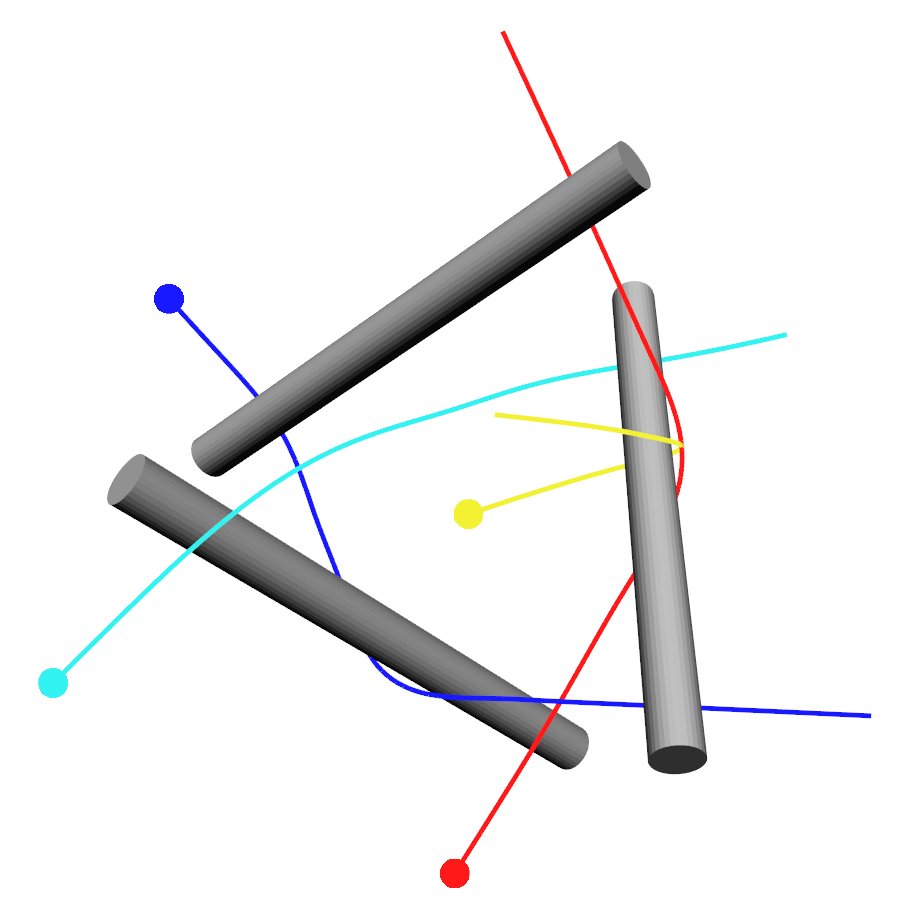}}
\stackinset{c}{24pt}{c}{-4pt}{\little}{\big}
\put(-518,26){(a)}
\put(-312,80){(b)}
\put(-540,140){ADMM}
\put(-540,70){Penalty-Method \cite{zhou2020ego2}}
\put(-25,30){(c)}
\vspace{-10px}
}
\hfill
\captionof{figure}{\small{Multiple UAV in complex environments, both with two groups of UAVs switching positions without (a) and with (b) obstacles. \revised{(c): The distributions of computational time of ADMM \ref{alg:ADMM1} and Penalty-Method \cite{zhou2020ego2} over $600$ trajectories computed for a synthetic environment in which we randomly generate initial trajectories for the $4$ UAVs in different homotopy classes.}}}
\label{fig:multiUAV}
\vspace{-15px}
\end{figure*}
\TE{Fast Separating Plane Update:} We found that iterative separating plane updates is a major computational bottleneck. Fortunately, mature collision detection algorithms such as the Gilbert–Johnson–Keerthi (GJK) algorithm \cite{montanari2017improving} can quickly return the optimal separating direction. We emphasize that our separating planes minimizing the soft log-barrier penalties do not match the separating directions returned by GJK in general. However, we found that the separating plane return by GJK oftentimes leads to a reduction in the Lagrangian function, while being orders of magnitude faster to compute due to their highly optimized implementation. \revised{Therefore, we propose to use the GJK algorithm for updating the separating planes. After each iteration of of either AM \ref{alg:AM} or ADMM \ref{alg:ADMM1}, we check whether the Lagrangian function $\mathcal{L}$ is decreasing. If $\mathcal{L}$ increases, we fallback to our standard log-barrier functions so the overall algorithm conforms to our convergence guarantee.}

\begin{table}[ht]
\centering
\footnotesize
\setlength{\tabcolsep}{5pt}
\begin{tabular}{ababa}
\hline
\multicolumn{5}{c}{Trajectory Length$(m)$ / Flying Time$(s)$}\\
\hline
Example & AM & ADMM & P-IPM \cite{2010.09904} & Safe-Corridor \cite{han2021fast}\\
\hline
(a)     & 16.0/9.6            & 16.0/9.6               & 16.1/9.7           & 18.0/14.0   \\
(b)     & 17.9/10.4            & 17.9/10.4               & 18.1/10.6           & 19.0/11.3   \\
(c)     & 13.2/8.9            & 13.2/9.0               & 13.8/9.1           & 14.6/11.7   \\
(d)     & 39.1/21.8            & 39.2/21.9               & 39.2/21.8           & 47.2/25.1   \\
(e)     & 66.8/44.7            & 66.5/44.6               & 68.1/45.5           & 71.2/57.9  \\
(f)     & 67.3/62.8           & 67.2/62.8              & 70.1/65.3          & 74.6/74.1   \\
\hline
\multicolumn{5}{c}{Computational Cost$(ms)$}\\
\hline
Example & AM & ADMM & P-IPM \cite{2010.09904} & Safe-Corridor \cite{han2021fast}\\
(a)     & 388                & 45                  & 1.0K               & 41   \\
(b)     & 133                & 34                  & 2.8K               & 50   \\
(c)     & 313                & 123                  & 7.8K               & 79   \\
(d)     & 8.7K                & 938                  & 34.3K               & 51   \\
(e)     & 39.0K                & 6.4K               & 175.0K             & 864   \\
(f)     & 92.9K              & 18.9K               & 420.1K             & 5.6K   \\
\hline
\end{tabular}
\caption{\label{table:qualitySingleUAV} \small{The quality of results in terms of trajectory length/flying time and computational cost of the three methods (ours, P-IPM \cite{2010.09904}, and Safe-Corridor \cite{han2021fast}) for the six examples of \prettyref{fig:singleUAV}.}}
\vspace{-15px}
\end{table}

\TE{Single-UAV:} We first show six examples of trajectory optimization for a single UAV in complex environments as illustrated in \prettyref{fig:singleUAV}. For each example, we compare our method against two baselines \cite{2010.09904,han2021fast}. We initialize all three methods using the same feasible trajectory that is manually designed. Our AM \ref{alg:AM} takes from $133$ to $93K(ms)$ to convergence and our ADMM \ref{alg:ADMM1} takes $34$ to $18.9K(ms)$, as compared with our prior work \cite{2010.09904} taking $1K$ to $420K(ms)$. The convergence history for three of the algorithms are summarized in \prettyref{fig:UAVConv}. Although the convergence speed is comparable to P-IPM in terms of number of iterations, our two methods (AM and ADMM) clearly outperform in terms of computational time. By not restricting the trajectory to precomputed corridors as done in \cite{7839930,8462878,tordesillas2019faster,han2021fast,wang2021geometrically}, our method allows a larger solution space and returns a shorter trajectory. We summarize the quality of trajectory as computed by three methods in \prettyref{table:qualitySingleUAV}. \revised{Finally, we conduct large-scale experiments using a synthetic problem illustrated in \prettyref{fig:singleUAV}g, where we randomly compute feasible initial trajectories using RRT-connect for $600$ times and compare the computational speed of ADMM \ref{alg:ADMM1} and \cite{2010.09904,han2021fast}. The resulting plot \prettyref{fig:singleUAV}g shows more than an order of magnitude's speedup over P-IPM using stiffness decoupling, with our trajectories having similar quality (mean/variance trajectory length 14.7/\num{2e-2} using ADMM versus 15.0/\num{4e-2} using P-IPM, and mean/varance flying time 12.7/\num{2e-5} using ADMM versus 12.8/\num{4e-5} using P-IPM). Although the computational speed of Safe-Corridor \cite{han2021fast} is slightly faster, our results give shorter trajectories (mean/variance trajectory length 15.9/\num{8e-2} and mean/variance flying time 14.8/\num{3e-2} using \cite{han2021fast}).}

\TE{Multi-UAV:} We assume the trajectory of each UAV is represented by composite B\'ezier curves with degree $M=5$. We use a bounding-volume hierarchy and only add collision constraints when the distance between two convex hulls are less than the activation distance of log-barrier function (0.1$m$). Note that this abrupt change in number of collision constraints will not hinder the convergence of ADMM because it can happen at most finitely many times. We compute initial trajectories using RRT connect. Our ADMM algorithm minimizes the jerk of each trajectory with time optimality as our objective function $O(\bar{X}_i,\Delta \bar{t}_i)$. \prettyref{fig:multiUAV} shows two challenging problems. We compare our method with penalty method \cite{zhou2020ego2}, which uses soft penalty terms to push trajectories out of the obstacles. This work is complementary to our method, which allows initial guesses to penetrate obstacles but cannot ensure final result to be collision-free. Instead, our method must start from a collision-free initial guess and maintain the collision-free guarantee throughout the optimization. In \prettyref{table:qualityMultiUAV}, we compare the quality of solutions and computational cost of these two methods. \revised{We have also conducted a large-scale comparison with penalty method \cite{zhou2020ego2} as illustrated in \prettyref{fig:multiUAV}c, where we randomly generate $600$ initial trajectories that cover different homotopy classes. As summarized in \prettyref{fig:multiUAV}c, our method generates trajectories with faster computing time and a $100\%$ success rate, while \cite{zhou2020ego2} can only achieve a success rate of $53\%$. Our results have higher quality than \cite{zhou2020ego2}, e.g. the mean/variance of trajectory length 77.8/17.8 (ours) vs 80.1/21.3 (\cite{zhou2020ego2}) and flying time 11.3/0.4 (ours) vs 20.6/8.0 (\cite{zhou2020ego2}).}

\begin{figure}[ht]
\centering
\scalebox{.7}{
\includegraphics[width=.24\textwidth]{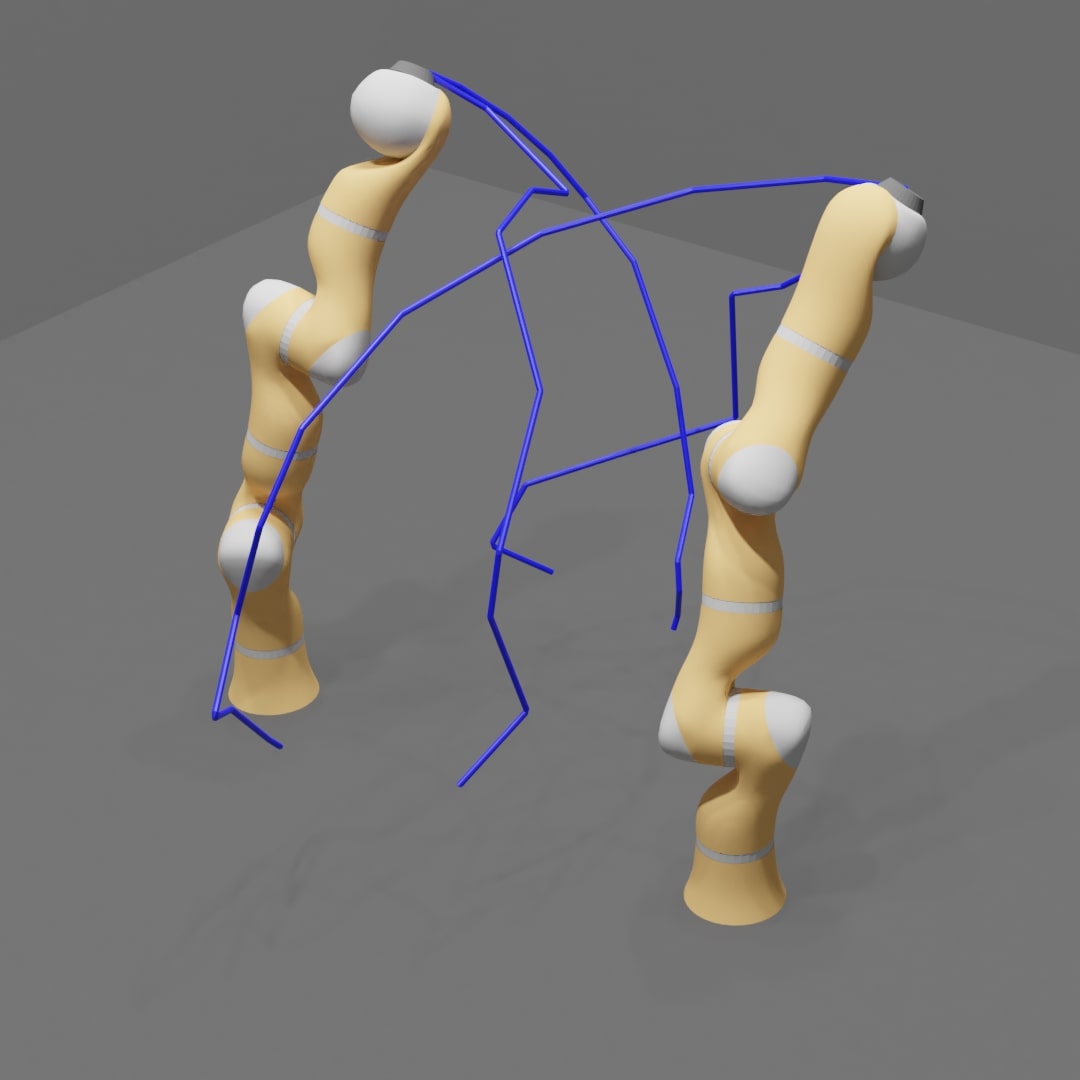}
\includegraphics[width=.24\textwidth]{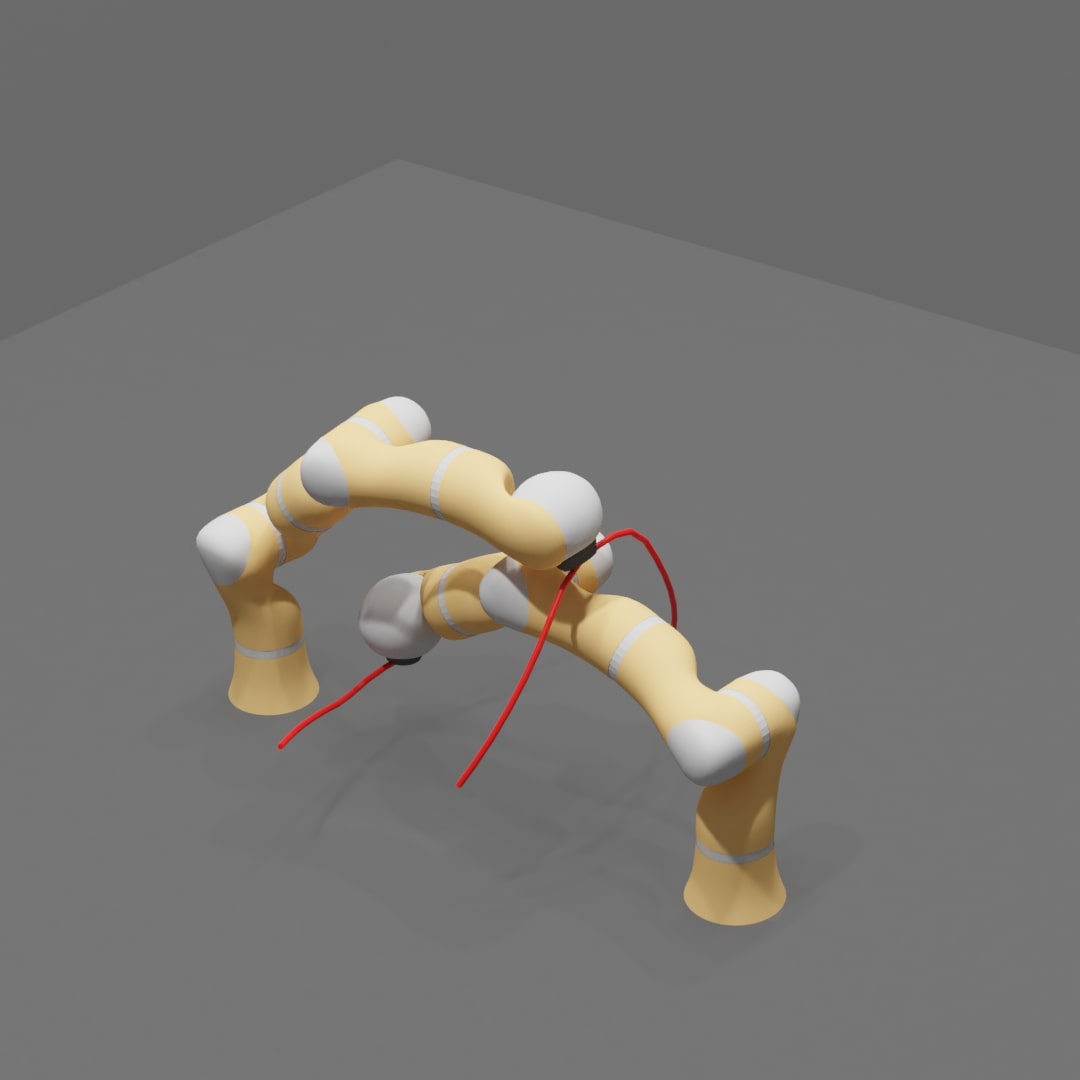}
\put(-150,10){(a)}
\put(-30,10){(b)}}
\caption{\label{fig:ARMBench}\small{Trajectory optimization for two KUKA LWR robot arms switching end-effector positions. (a): initial trajectory via RRT-connect; (b): optimized trajectory.}}
\vspace{-15px}
\end{figure}

\begin{figure}[ht]
\centering
\scalebox{.95}{
\includegraphics[width=.48\textwidth]{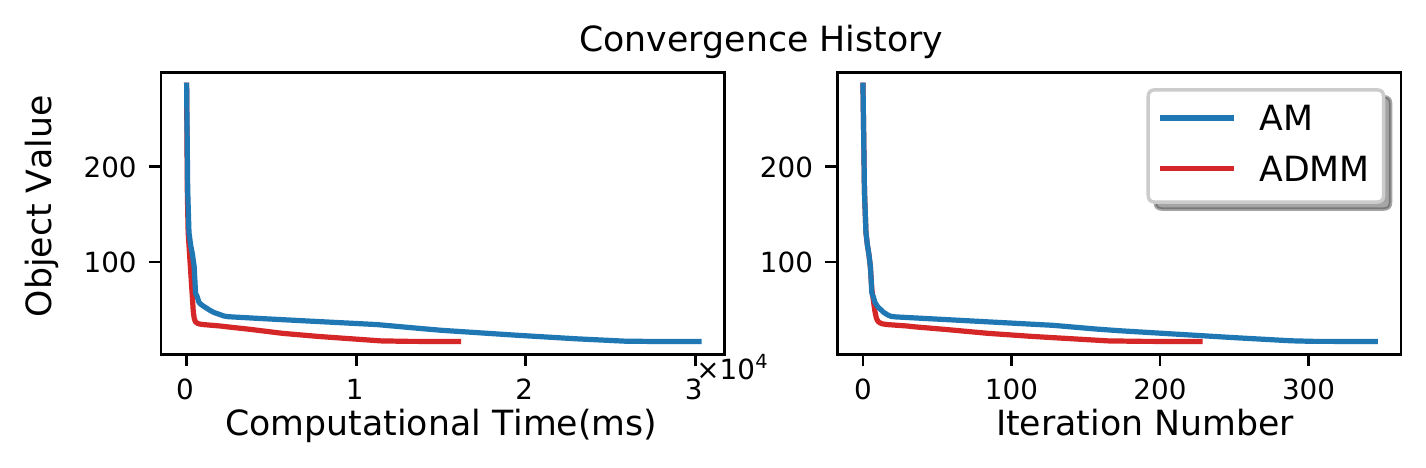}
\put(-150,30){(a)}
\put(-30,30){(b)}}
\vspace{-5px}
\caption{\label{fig:armConv}\small{Convergence history for the example in \prettyref{fig:ARMBench}, comparing AM and stiffness-decoupled ADMM in terms of wall-time (a) and \#Iterations (b).}}
\vspace{-15px}
\end{figure}

\begin{table}[ht]
\centering
\setlength{\tabcolsep}{5pt}
\begin{tabular}{aba}
\hline
\multicolumn{3}{c}{Trajectory Length$(m)$ / Flying Time$(s)$ / Computational Cost(ms)}\\
\hline
Example & ADMM & Penalty-Method \cite{zhou2020ego2} \\
\hline
(a)     & 152.68/7.84/515            & 169.34/14.48/844.49    \\
(b)     & 71.33/14.57/2527            & 89.82/20.73/584.89    \\
\hline
\end{tabular}
\caption{\label{table:qualityMultiUAV} \small{The quality of results in terms of trajectory length/flying time and computational cost, comparing our method and Penalty-Method \cite{zhou2020ego2}.}}
\vspace{-8px}
\end{table}
\TE{Articulated Robot Arm:} We highlight the performance of our method via an example involving two arms. We approximate each robot link as a single convex object to reduce the number of separating planes. Our example is inspired by prior work \cite{ha2020multiarm}, as illustrated in \prettyref{fig:ARMBench}, where we have two KUKA LWR robot arms (each with $8$ links) switch positions of their end-effectors. From an initial trajectory computed with the length 6.38m using RRT-connect, our stiffness-decoupled ADMM method can easily minimize acceleration of end-effects obtaining a trajectory with the length 1.63m in 16s. The convergence history is shown in \prettyref{fig:armConv}. 


%% file: conclusion.tex
\section{Conclusion \& Limitation\label{sec:conclusion}}
We propose a variant of ADMM-type solver for trajectory \revised{optimization}. We observe that the limited efficacy of our prior work \cite{2010.09904} is mainly due to the stiff log-barrier functions corresponding to various hard constraints. Therefore, we propose to decompose stiff and non-stiff objective function terms using slack variables, while using additional constraints to ensure their consistency. ADMM was  originally applied to convex optimizations and we establish its convergence guarantee under non-convex objectives and constraints that arise from UAV and articulated robot kinematics. Our experiments confirm that ADMM successfully resolves stiff-coupling issues and achieves tens of times' speedup over Newton-type algorithms. \revised{The major limitation of our method is the requirement of a strictly feasible initial trajectory and our convergence rate can also be dependent on the initial guess. As a result, our method cannot be used for receding-horizon settings. If the horizons are truncated, then there can be unforeseen obstacles, with which collisions are ignored, resulting in infeasible trajectories in future horizons. Finally, the convergence of ADMM-type solvers for general control of nonlinear dynamic systems \cite{10.1145/3016963} remains an open question.}

%% file: convergenceAM.tex
\section{\label{sec:convAM}Convergence Analysis:\\ Alternating Minimization (AM)}
Throughout our analysis, we assume the objective function $\mathcal{O}$ is twice-differentiable. We rewrite the Lagrangian \prettyref{eq:prob4Log} as a function in the three (sets of) variables, $\Theta,P_i,P_{ij}$, where we define the shorthand notation $\Theta\triangleq\TWO{\theta}{\Delta t}$, $P_i\triangleq\TWO{n_i}{d_i}$, and $P_{ij}\triangleq\TWO{n_{ij}}{d_{ij}}$. We also denote $P\triangleq\TWO{n}{d}$ as an arbitrary plane, which can be some $P_i$ or $P_{ij}$. During the $k$th iteration, these three variables will be updated in order and \prettyref{alg:AM} will generate an infinite sequence $\{\THREE{\Theta^k}{P_i^k}{P_{ij}^k}\}$. We further assume that each minimization subproblem of \prettyref{alg:AM}, e.g., the $\Theta^k$-subproblem, is warm-started from its previous value $\Theta^{k-1}$. The convergence of AM in non-convex settings has been shown in prior work \cite{xu2017globally}, but they assume the log-barrier functions are proximable, which does not hold in our case due to the non-linear forward kinematics $x(\theta)$. Other work such as \cite{deng2019efficiency} assumes the availability of a global Lipschitz constant, which does not hold for a log-barrier function. We overcome this difficulty by using a line-search to ensure the satisfaction of Armijo's condition for solving the three subproblems. In other words, \prettyref{ln:AMProb1} of \prettyref{alg:AM} is implemented in \prettyref{alg:AMProb1}. To implement \prettyref{ln:AMProb2} and \prettyref{ln:AMProb3} of \prettyref{alg:AM}, we handle the additional unit-norm constraint on $n_i$ and $n_{ij}$ via Riemannian optimization \cite{ring2012optimization}. Taking an arbitrary $P$ for example, we start from a feasible initial guess $\|n\|=1$ and optimize on the tangent space $v\in\mathcal{T}_nS^2$. The optimized $v$ provides an updated $P=\TWO{\exp_nv}{d}$. Essentially, we use standard line-search as outlined in \prettyref{alg:AMProb2} to optimize the function $\mathcal{L}(\exp_nv,d)$ denoted as $\mathcal{L}\circ\exp_n$. We further use a shorthand notation $\mathcal{L}\circ\exp(\theta,\Delta t,v_i,d_i,v_{ij},d_{ij})$ to denote the reparameterization of $\mathcal{L}$ with each $n_i$ (w.r.t. $n_{ij}$) replaced by $\exp_{n_i}(v_i)$ (w.r.t. $\exp_{n_{ij}}(v_{ij})$). A direct verification shows that a first-order critical point of $\mathcal{L}$ is a point with $\|\nabla\mathcal{L}\circ\exp\|=0$.
\begin{algorithm}[ht]
\caption{\label{alg:AMProb1} Line-Search Gradient Descent}
\small{\begin{algorithmic}[1]
\Require{Initial guess $\Theta$, $c\in(0,1)$, $\gamma\in(0,1)$}
\State $\alpha\gets1$
\While{$\mathcal{L}(\Theta-\alpha\nabla_\Theta\mathcal{L}(\Theta))>
\mathcal{L}(\Theta)-c\alpha\|\nabla_\Theta\mathcal{L}(\Theta)\|^2$}
\State $\alpha\gets\gamma\alpha$
\EndWhile
\State Return $\Theta-\alpha\nabla_\Theta\mathcal{L}(\Theta)$
\end{algorithmic}}
\end{algorithm}
\begin{algorithm}[ht]
\caption{\label{alg:AMProb2} Riemannian Line-Search Gradient Descent}
\small{\begin{algorithmic}[1]
\Require{Initial guess $v=0$, $c\in(0,1)$, $\gamma\in(0,1)$}
\State $\alpha\gets1$
\While{$\substack{\mathcal{L}\circ\exp_n(\TWO{v}{d}-\alpha\nabla_{v,d}\mathcal{L}\circ\exp_n)\\>
\mathcal{L}\TWO{v}{d}-c\alpha\|\nabla_{v,d}\mathcal{L}\circ\exp_n\|^2}$}
\State $\alpha\gets\gamma\alpha$
\EndWhile
\State $\TWO{v}{d}\gets \TWO{v}{d}-\alpha\nabla_{v,d}\mathcal{L}\circ\exp_n\TWO{v}{d}$
\State Return $P_i\gets\TWO{\exp_nv}{d}$
\end{algorithmic}}
\end{algorithm}

To establish convergence to first-order critical point, we bound the magnitude of $\alpha$ making use of the Lipschitz constant of a function. In the following analysis, we omit some parameters of $\mathcal{L}$ for brevity, where the omitted parameters is same as \prettyref{alg:AM}.
\begin{lemma}
\label{lem:AM1}
Consider a function $\mathcal{L}(\Theta)$. Assume that $\nabla_\Theta\mathcal{L}(\Theta)$ is $L$-Lipschitz continuous within some convex set $B$. If \prettyref{alg:AMProb1} updates $\Theta^k$ to $\Theta^{k+1}$ with $\Theta^k,\Theta^{k+1}\in B$, then we have: $\alpha\geq2\gamma(1-c)/L$ and $\mathcal{L}(\Theta^{k+1})+c\|\Theta^{k+1}-\Theta^k\|^2\leq\mathcal{L}(\Theta^k)$.
\end{lemma}
\begin{IEEEproof}
When $\alpha\leq2(1-c)/L$, we have:
\small
\begin{align*}
\mathcal{L}(\Theta^{k+1})
\leq&\mathcal{L}(\Theta^k)+\langle\nabla_\Theta\mathcal{L}(\Theta^k),\Theta^{k+1}-\Theta^k\rangle+\frac{L}{2}\|\Theta^{k+1}-\Theta^k\|^2\\
=&\mathcal{L}(\Theta^k)+(\frac{L\alpha^2}{2}-\alpha)\|\nabla_\Theta\mathcal{L}(\Theta^k)\|^2\\
\leq&\mathcal{L}(\Theta^k)-c\alpha\|\nabla_\Theta\mathcal{L}(\Theta^k)\|^2.
\end{align*}
\normalsize
By the logic of \prettyref{alg:AMProb1}, we have the first inequality:
\begin{align*}
\alpha\geq\gamma^{\lceil\log_\gamma(2(1-c)/L)\rceil}\geq2\gamma(1-c)/L.\labelthis{eq:AMProb1Bound1}
\end{align*}
The second inequality follows from the Armijo's condition:
\small
\begin{align*}
&\mathcal{L}(\Theta^{k+1})+c\|\Theta^{k+1}-\Theta^k\|^2=\mathcal{L}(\Theta^{k+1})+c\alpha^2\|\nabla_\Theta\mathcal{L}(\Theta^k)\|^2\\
\leq&\mathcal{L}(\Theta^k)-c\alpha\|\nabla_\Theta\mathcal{L}(\Theta^k)\|^2+c\alpha^2\|\nabla_\Theta\mathcal{L}(\Theta^k)\|^2\leq\mathcal{L}(\Theta^k),\labelthis{eq:AMProb1Bound2}
\end{align*}
\normalsize
where we have used $\alpha\leq1$.
\end{IEEEproof}
Given the above results, it is straightforward to establish the convergence guarantee for a convergent sub-sequence.
\begin{lemma}
\label{lem:AM2}
If $\mathcal{L}$ is lower-bounded, then every accumulation point of the sequence $\{\THREE{\Theta^k}{P_i^k}{P_{ij}^k}\}$ generated by \prettyref{alg:AM} is a critical point. Suppose the sequence is convergent, the local convergence speed of $\|\nabla\mathcal{L}\circ\exp\|$ is $\mathcal{O}(1/\sqrt{K})$.
\end{lemma}
\begin{IEEEproof}
The function $\{\mathcal{L}(\Theta^k,P_i^k,P_{ij}^k)\}$ is monotonically decreasing and bounded from below, so it converges to $\mathcal{L}(\bar{\Theta},\bar{P}_i,\bar{P}_{ij})$ by continuity, where $\THREE{\bar{\Theta}}{\bar{P}_i}{\bar{P}_{ij}}$ is an accumulation point with the convergent sub-sequence identified by the index subset: $\mathcal{K}\subset\{k|k=1,2,\cdots\}$.

\TE{Applying \prettyref{lem:AM1}:} Invoke \prettyref{eq:AMProb1Bound2} for $\mathcal{L}$ in \prettyref{alg:AMProb1} and we will have:
\begin{align*}
\mathcal{L}(\Theta^{k+1})+c\|\Theta^{k+1}-\Theta^k\|^2\leq\mathcal{L}(\Theta^k).
\end{align*}
Invoke \prettyref{eq:AMProb1Bound2} for $\mathcal{L}\circ\exp_n$ in \prettyref{alg:AMProb2} and we will have:
\begin{align*}
\mathcal{L}(P^{k+1})+c\|v^k\|^2+c\|d^{k+1}-d^k\|^2\leq\mathcal{L}(P^k).
\end{align*}
Summing up the above equations for all the iterations of \prettyref{alg:AM} and we have:
\small
\begin{align*}
&\sum_{k=0}^\infty\|\Theta^{k+1}-\Theta^k\|^2+\\
&\sum_{k=0}^\infty\sum_i\|\TWO{v_i^k}{d_i^{k+1}}-\TWO{0}{d_i^k}\|^2+\\
&\sum_{k=0}^\infty\sum_{ij}\|\TWO{v_{ij}^k}{d_{ij}^{k+1}}-\TWO{0}{d_{ij}^k}\|^2\\
\leq&\frac{\mathcal{L}(\Theta^0,P_i^0,P_{ij}^0)-\mathcal{L}(\bar{\Theta},\bar{P}_i,\bar{P}_{ij})}{c}<\infty.
\end{align*}
\normalsize
We conclude that $\lim_{k\to\infty}\|\Theta^{k+1}-\Theta^k\|=0$, $\lim_{k\to\infty}\|v^k\|=0$, and $\lim_{k\to\infty}\|d^{k+1}-d^k\|=0$ for any $P$. Here we use shorthand notation $v_i^k$ (w.r.t. $v_{ij}^k$) to denote $v$ found by \prettyref{alg:AMProb2} for $P_i$ (w.r.t. $P_{ij}$). 

\TE{Optimality \& Convergence Speed of $\bar{\Theta}$:} $\DOM\mathcal{L}$ is dictated by the log-barrier function, which is an open set. $\mathcal{L}(\bar{\Theta},\bar{P}_i,\bar{P}_{ij})<\infty$ so that $\THREE{\bar{\Theta}}{\bar{P}_i}{\bar{P}_{ij}}$ is within the open set. As a result, we can find some small neighborhood $B_\rho\THREE{\bar{\Theta}}{\bar{P}_i}{\bar{P}_{ij}}\subset\DOM\mathcal{L}$ (denoted as $B_\rho$ for short) such that $\THREE{\Theta^k}{P_i^k}{P_{ij}^k}\in B_{\rho/2}$ and $\|\Theta^{k+1}-\Theta^k\|\leq\rho/2$ for any sufficiently large $\mathcal{K}\ni k>k_0$, where $\rho$ is some small positive number. W.l.o.g., we can discard the sequence before $k_0$ and assume $k_0=0$. We can choose a sufficiently small $\rho$ such that $\nabla_\Theta\mathcal{L}$ is $L$-Lipschitz continuous within $B_\rho\subset\DOM\mathcal{L}$.  Invoke \prettyref{eq:AMProb1Bound1} for $\mathcal{L}$ in \prettyref{alg:AMProb1} and we will have:
\small
\begin{align*}
0=&\lim_{\mathcal{K}\ni k\to\infty}\|\Theta^{k+1}-\Theta^k\|=\lim_{\mathcal{K}\ni k\to\infty}\alpha^k\|\nabla_\Theta\mathcal{L}(\Theta^k)\|\\
\geq&2\gamma(1-c)/L\lim_{\mathcal{K}\ni k\to\infty}\|\nabla_\Theta\mathcal{L}(\Theta^k)\|=2\gamma(1-c)/L\|\nabla_\Theta\mathcal{L}(\bar{\Theta})\|,
\end{align*}
\normalsize
where $\alpha^k$ is the $\alpha$ used in the line-search during the $k-$th iteration. If the entire sequence is convergent, then we have the following result for the local convergence speed:
\begin{align*}
&K(2\gamma(1-c)/L)^2\min_{k=0,\cdots,K}\|\nabla_\Theta\mathcal{L}(\Theta^k)\|^2\\
\leq&(2\gamma(1-c)/L)^2\sum_{k=0}^K\|\nabla_\Theta\mathcal{L}(\Theta^k)\|^2\leq\sum_{k=0}^K\|\Theta^{k+1}-\Theta^k\|^2\\
\leq&\sum_{k=0}^\infty\|\Theta^{k+1}-\Theta^k\|^2<\infty.
\end{align*}

\TE{Optimality \& Convergence Speed of $\bar{P}$:} We can also choose sufficiently small $\rho$ such that for any fixed $\THREE{\Theta^k}{P_i^k}{P_{ij}^k}\in B_\rho\subset\DOM\mathcal{L}$, the function $\nabla_{v,d}\mathcal{L}\circ\exp_n$ is $L$-Lipschitz continuous within $B_\rho$. We invoke \prettyref{eq:AMProb1Bound1} for $\mathcal{L}\circ\exp$ in \prettyref{alg:AMProb1} and following the same logic as that of $\bar{\Theta}$.
\end{IEEEproof}
Note that we cannot have overall convergence speed due to a lack of a global Lipschitz constant. To get rid of the whole sequence convergence assumption and establish a local convergence speed and the convergence of entire sequence $\{\THREE{\Theta^k}{P_i^k}{P_{ij}^k}\}$, we need the K{\L} property \cite{lojasiewicz1963topological} of $\mathcal{L}$, which is a rather mild assumption that holds for all real analytic functions. Specifically, we assume the K{\L} property \cite{lojasiewicz1963topological} holds for the composite function $\mathcal{L}\circ\exp$ defined below:
\begin{definition}
\label{def:KL}
The function $\mathcal{L}\circ\exp$ satisfies the K{\L} property at some point $\THREE{\bar{\Theta}}{\bar{P}_i}{\bar{P}_{ij}}$, if there exists a neighborhood $B_\rho$ and a convex function $\phi$ such that, for any $\THREE{\Theta}{\bar{P}_i}{\bar{P}_{ij}}\in B_\rho$, we have:
\begin{align*}
&\phi'(\mathcal{L}\THREE{\Theta}{P_i}{P_{ij}}-\mathcal{L}\THREE{\bar{\Theta}}{\bar{P}_i}{\bar{P}_{ij}})\\
&\|\nabla\mathcal{L}\FIVE{\Theta}{\exp_{n_i}(0)}{d_i}{\exp_{n_{ij}}(0)}{d_{ij}}\|\geq1.
\end{align*}
\end{definition}
We establish the convergence of sequence $\{\THREE{\Theta^k}{P_i^k}{P_{ij}^k}\}$ and the local convergence speed assuming the K{\L} property.
\begin{lemma}
\label{lem:AM3}
If $\mathcal{L}\circ\exp$ is lower-bounded and pertains the K{\L} property (\prettyref{def:KL}), and the sequence $\{\THREE{\Theta^k}{P_i^k}{P_{ij}^k}\}$ generated by \prettyref{alg:AM} has a finite accumulation point, then the sequence is convergent.
\end{lemma}
\begin{IEEEproof}
We denote $\THREE{\bar{\Theta}}{\bar{P}_i}{\bar{P}_{ij}}$ as the accumulation point. By invoking \prettyref{lem:AM2}, we know that $\nabla\mathcal{L}\circ\exp=0$. W.l.o.g., we assume $\mathcal{L}\THREE{\bar{\Theta}}{\bar{P}_i}{\bar{P}_{ij}}=0$. By continuity, we can choose a small neighborhood $B_{2\rho}\subset\DOM\mathcal{L}$ where $\|\nabla\mathcal{L}\circ\exp\|<\epsilon$ and the the K{\L} property holds. We further denote $\nabla\mathcal{L}\circ\exp$ as $L$-Lipschitz continuous within $B_{2\rho}$. Finally, in a sufficiently small local neighborhood, we can setup strong equivalence between metrics in tangent space and ambient space. In other words, we can find two positive constants $\underline{L}$ and $\bar{L}$ such that:
\begin{align*}
\underline{L}\|\exp_nv-\exp_nv'\|\leq\|v-v'\|\leq\bar{L}\|\exp_nv-\exp_nv'\|,
\end{align*}
for any point in $B_{2\rho}$. If we start from $\THREE{\Theta^k}{P_i^k}{P_{ij}^k}\in B_\rho$, we choose sufficiently large $k_0$ such that the subsequent point $\THREE{\Theta^{k+1}}{P_i^{k+1}}{P_{ij}^{k+1}}\in B_{2\rho}$ for every $k\geq k_0$ due to \prettyref{lem:AM2}. W.l.o.g., we can assume $k_0=0$. Let's now define two constants:
\begin{align*}
L_1\triangleq&\max(\frac{L}{2\gamma(1-c)}+\sum_iL+\sum_{ij}L,\frac{L\bar{L}}{2\gamma(1-c)})\\
L_2\triangleq&\min(c,c{\underline{L}}^2).
\end{align*}
Since there is a sub-sequence converging to $\THREE{\bar{\Theta}}{\bar{P}_i}{\bar{P}_{ij}}$, we can choose large enough $k\geq k_0$ such that:
\footnotesize
\begin{align*}
\|\THREE{\Theta^k}{P_i^k}{P_{ij}^k}-\THREE{\bar{\Theta}}{\bar{P}_i}{\bar{P}_{ij}}\|
\leq\rho-L_1/L_2\phi(\mathcal{L}\THREE{\Theta^k}{P_i^k}{P_{ij}^k}).
\end{align*}
\normalsize
W.l.o.g., we can assume $k_0=0$ and:
\begin{align}
\label{eq:KL0}
\THREE{\Theta^0}{P_i^0}{P_{ij}^0}\in B_{\rho-L_1/L_2\phi(\mathcal{L}\THREE{\Theta^0}{P_i^0}{P_{ij}^0})}.
\end{align}
Next, we show that every $\THREE{\Theta^k}{P_i^k}{P_{ij}^k}\in B_\rho$ by induction. If we already have $\THREE{\Theta^0}{P_i^0}{P_{ij}^0},\cdots,\THREE{\Theta^k}{P_i^k}{P_{ij}^k}\in B_\rho$, then the following holds for $\THREE{\Theta^{k+1}}{P_i^{k+1}}{P_{ij}^{k+1}}\in B_{\rho}$:
\footnotesize
\begin{align*}
&\frac{\mathcal{L}\THREE{\Theta^k}{P_i^k}{P_{ij}^k}-\mathcal{L}\THREE{\Theta^{k+1}}{P_i^{k+1}}{P_{ij}^{k+1}}}
{\phi(\mathcal{L}\THREE{\Theta^k}{P_i^k}{P_{ij}^k})-\phi(\mathcal{L}\THREE{\Theta^{k+1}}{P_i^{k+1}}{P_{ij}^{k+1}})}\\
\leq&1/\phi'(\mathcal{L}\THREE{\Theta^k}{P_i^k}{P_{ij}^k})\\
\leq&\|\nabla\mathcal{L}(\Theta^k,\exp_{n_i^k}(0),d_i^k,\exp_{n_{ij}^k}(0),d_{ij}^k)\|\\
=&\|\nabla_\Theta\mathcal{L}(\Theta^k)\|+\\
&\sum_i\|\nabla_{v,d}\mathcal{L}\circ\exp_{n_i^k}(0,d_i^k)\|+
\sum_{ij}\|\nabla_{v,d}\mathcal{L}\circ\exp_{n_{ij}^k}(0,d_{ij}^k))\|\\
\leq&\frac{\|\Theta^{k+1}-\Theta^k\|}{\alpha_\Theta^k}+\\
&\sum_i\left[\frac{\|\TWO{v_i^k}{d_i^{k+1}-d_i^k}\|}{\alpha_{n_i,d_i}^k}+L\|\Theta^{k+1}-\Theta^k\|\right]+\\
&\sum_{ij}\left[\frac{\|\TWO{v_{ij}^k}{d_{ij}^{k+1}-d_{ij}^k}\|}{\alpha_{n_{ij},d_{ij}}^k}+L\|\Theta^{k+1}-\Theta^k\|\right]\\
\leq&\left[\frac{L}{2\gamma(1-c)}+\sum_iL+\sum_{ij}L\right]\|\Theta^{k+1}-\Theta^k\|+\\
&\frac{L}{2\gamma(1-c)}\left[\sum_i\|\TWO{v_i^k}{d_i^{k+1}-d_i^k}\|+\sum_{ij}\|\TWO{v_{ij}^k}{d_{ij}^{k+1}-d_{ij}^k}\|\right]\\
\leq&\left[\frac{L}{2\gamma(1-c)}+\sum_iL+\sum_{ij}L\right]\|\Theta^{k+1}-\Theta^k\|+\\
&\frac{L\bar{L}}{2\gamma(1-c)}\left[\sum_i\|P_i^{k+1}-P_i^k\|+\sum_{ij}\|P_{ij}^{k+1}-P_{ij}^k\|\right]\\
\leq&L_1\|\THREE{\Theta^{k+1}}{P_i^{k+1}}{P_{ij}^{k+1}}-\THREE{\Theta^k}{P_i^k}{P_{ij}^k}\|,\labelthis{eq:KL1}
\end{align*}
\normalsize
where we use shorthand notation $\alpha_\Theta^k$ (w.r.t. $\alpha_{n_i,d_i}^k,\alpha_{n_{ij},d_{ij}}^k$) to denote $\alpha$ found by \prettyref{alg:AMProb1} for $\Theta$ (w.r.t. $P_i,P_{ij}$). The first inequality above is due to the convexity of $\phi$. The second inequality is due to the K{\L} property of $\mathcal{L}\circ\exp$. The third inequality is due to the logic of line-search algorithms and $L$-Lipschitz continuity of gradient within $B_{2\rho}$. The forth inequality is due to \prettyref{lem:AM2} and the fifth inequality is due to the strong metric equivalently, both within $B_{2\rho}$. We further have the following estimate of function decrease due to \prettyref{lem:AM2}:
\small
\begin{align*}
&\mathcal{L}\THREE{\Theta^k}{P_i^k}{P_{ij}^k}-\mathcal{L}\THREE{\Theta^{k+1}}{P_i^{k+1}}{P_{ij}^{k+1}}\\
\geq&c\|\Theta^{k+1}-\Theta^k\|^2+c\sum_i\|v_i^k\|^2+c\sum_i\|d_i^{k+1}-d_i^k\|^2+\\
&c\sum_{ij}\|v_{ij}^k\|^2+c\sum_{ij}\|d_{ij}^{k+1}-d_{ij}^k\|^2\\
\geq&c\|\Theta^{k+1}-\Theta^k\|^2+c\underline{L}^2\sum_i\|n_i^{k+1}-n_i^k\|^2+c\sum_i\|d_i^{k+1}-d_i^k\|^2+\\
&c\underline{L}^2\sum_{ij}\|n_{ij}^{k+1}-n_{ij}^k\|^2+c\sum_{ij}\|d_{ij}^{k+1}-d_{ij}^k\|^2\\
\geq&L_2\|\THREE{\Theta^{k+1}}{P_i^{k+1}}{P_{ij}^{k+1}}-\THREE{\Theta^k}{P_i^k}{P_{ij}^k}\|^2.\labelthis{eq:KL2}
\end{align*}
\normalsize
Combining \prettyref{eq:KL1} and \prettyref{eq:KL2}, we conclude that:
\begin{align*}
&\phi(\mathcal{L}\THREE{\Theta^k}{P_i^k}{P_{ij}^k})-
\phi(\mathcal{L}\THREE{\Theta^{k+1}}{P_i^{k+1}}{P_{ij}^{k+1}})\\
\geq&L_2/L_1\|\THREE{\Theta^{k+1}}{P_i^{k+1}}{P_{ij}^{k+1}}-\THREE{\Theta^k}{P_i^k}{P_{ij}^k}\|.\labelthis{eq:KL3}
\end{align*}
The above property holds for all $0,\cdots,k$ so we can sum these equations up to derive:
\small
\begin{align*}
&\|\THREE{\Theta^{k+1}}{P_i^{k+1}}{P_{ij}^{k+1}}-\THREE{\bar{\Theta}}{\bar{P}_i}{\bar{P}_{ij}}\|\\
=&\|\THREE{\bar{\Theta}}{\bar{P}_i}{\bar{P}_{ij}}-\THREE{\Theta^0}{P_i^0}{P_{ij}^0}\|+\\
&\|\THREE{\Theta^{k+1}}{P_i^{k+1}}{P_{ij}^{k+1}}-\THREE{\Theta^0}{P_i^0}{P_{ij}^0}\|\\
\leq&\|\THREE{\bar{\Theta}}{\bar{P}_i}{\bar{P}_{ij}}-\THREE{\Theta^0}{P_i^0}{P_{ij}^0}\|+\\
&L_1/L_2\left[\phi(\mathcal{L}\THREE{\Theta^0}{P_i^0}{P_{ij}^0})-
\phi(\mathcal{L}\THREE{\Theta^{k+1}}{P_i^{k+1}}{P_{ij}^{k+1}})\right]\\
\leq&\rho-L_1/L_2\phi(\mathcal{L}\THREE{\Theta^0}{P_i^0}{P_{ij}^0})+\\
&L_1/L_2\left[\phi(\mathcal{L}\THREE{\Theta^0}{P_i^0}{P_{ij}^0})-
\phi(\mathcal{L}\THREE{\Theta^{k+1}}{P_i^{k+1}}{P_{ij}^{k+1}})\right]\leq\rho,
\end{align*}
\normalsize
where we have used \prettyref{eq:KL0} in the second to last inequality. By induction, we conclude that every element of $\THREE{\Theta^k}{P_i^k}{P_{ij}^k}\in B_\rho$ and we have from \prettyref{eq:KL3}:
\begin{align*}
\sum_{k=0}^\infty\|\THREE{\Theta^{k+1}}{P_i^{k+1}}{P_{ij}^{k+1}}-\THREE{\Theta^k}{P_i^k}{P_{ij}^k}\|<\infty,
\end{align*}
which shows that the sequence $\{\THREE{\Theta^k}{P_i^k}{P_{ij}^k}\}$ is a Cauchy sequence and is thus convergent. Since $\THREE{\bar{\Theta}}{\bar{P}_i}{\bar{P}_{ij}}$ is a limit point of the sequence, we conclude that the entire sequence converge to $\THREE{\bar{\Theta}}{\bar{P}_i}{\bar{P}_{ij}}$ with local convergence speed being $\mathcal{O}(1/\sqrt{K})$ due to \prettyref{lem:AM2}.
\end{IEEEproof}

%% file: convergenceADMM1.tex
\section{\label{sec:convADMM1}Convergence Analysis:\\ ADMM with Stiffness Decoupling}
We prove the convergence of \prettyref{alg:ADMM1} which is similar to \cite{jiang2019structured} but we deviate from their prove in that 1) we consider nonlinear constraints and 2) we consider a Lagrangian function (\prettyref{eq:ALF}) that is not Lipschitz continuous. To simplify notation, we present our proof without variable $\Delta t$ and $\Delta \bar{t}$, i.e., we assume $\Theta=\theta$ in the following proof. In fact, the identical proof can be extended to the case with time-optimality. Specifically, if we adopt the following change of variable, the case with time optimality is proved:
\small
\begin{align*}
\theta\gets\TWOC{\theta}{\Delta t}\quad
X_i(\theta)\gets\TWOC{X_i(\theta)}{\Delta t}\quad
\bar{X}_i\gets\TWOC{\bar{X}_i}{\Delta \bar{t}_i}\quad
\lambda_i\gets\TWOC{\lambda_i}{\Lambda_i}.
\end{align*}
\normalsize
We take the same assumptions on $\mathcal{O}$ as that of \prettyref{sec:convAM}. The only different from \prettyref{alg:AM} lies in \prettyref{ln:ADMMProb1} and \prettyref{ln:ADMMProb2} of \prettyref{alg:ADMM1}. First, it is not practical to solve the $\bar{X}_i^{k+1}$-subproblem exactly and we adopt the linear proximal operator instead:
\small
\begin{align*}
\bar{X}_i^{k+1}=&\argmin{\bar{X}_i}\frac{\beta}{2}\|\bar{X}_i-\bar{X}_i^k\|+\\
&\langle\nabla_{\bar{X}_i}\mathcal{L}(\theta^{k+1},\bar{X}_i,\lambda_i^k,n_i^k,d_i^k,n_{ij}^k,d_{ij}^k),\bar{X}_i-\bar{X}_i^k\rangle\\
=&\bar{X}_i^k-\frac{1}{\beta}\left[\nabla\mathcal{O}(\bar{X}_i^k)+\varrho(\bar{X}_i^k-X_i(\theta^{k+1}))-\lambda_i^k\right]\\
=&\bar{X}_i^k-\frac{1}{\beta}\left[\nabla\mathcal{O}(\bar{X}_i^k)-\varrho(\bar{X}_i^{k+1}-\bar{X}_i^k)-\lambda_i^{k+1}\right],\labelthis{eq:ADMM1}
\end{align*}
\normalsize
with $\beta$ determining the strength of regularization. The change of $\mathcal{L}$ due to \prettyref{ln:ADMMProb2} can be bounded as follows:
\begin{lemma}
\label{lem:ADMM1}
If $\nabla\mathcal{O}$ is $L_\mathcal{O}$-Lipschitz continuous, then \prettyref{ln:ADMMProb2} of \prettyref{alg:ADMM1} will monotonically increase $\mathcal{L}$ by at most:
\begin{align*}
\frac{2(\beta-\varrho)^2}{\varrho}\|\bar{X}_i^{k+1}-\bar{X}_i^k\|^2+\frac{2(\beta-\varrho+L_\mathcal{O})^2}{\varrho}\|\bar{X}_i^k-\bar{X}_i^{k-1}\|^2.
\end{align*}
\end{lemma}
\begin{IEEEproof}
Some minor rearrangement of \prettyref{eq:ADMM1} would lead to:
\begin{align*}
(\beta-\varrho)(\bar{X}_i^{k+1}-\bar{X}_i^k)=\lambda_i^{k+1}-\nabla\mathcal{O}(\bar{X}_i^k).\labelthis{eq:ADMM2}
\end{align*}
We have the following identity due to $L_\mathcal{O}$-Lipschitz property of $\nabla\mathcal{O}$:
\small
\begin{align*}
&\|\lambda_i^{k+1}-\lambda_i^k\|\leq\|\nabla\mathcal{O}(\bar{X}_i^k)-\nabla\mathcal{O}(\bar{X}_i^{k-1})\|+\\
&(\beta-\varrho)\|\bar{X}_i^{k+1}-\bar{X}_i^k\|+(\beta-\varrho)\|\bar{X}_i^k-\bar{X}_i^{k-1}\|\\
\leq&(\beta-\varrho)\|\bar{X}_i^{k+1}-\bar{X}_i^k\|+(\beta-\varrho+L_\mathcal{O})\|\bar{X}_i^k-\bar{X}_i^{k-1}\|.
\end{align*}
\normalsize
By the inequality $(a+b)^2\leq2a^2+2b^2$, we can derive the result to be proved.
\end{IEEEproof}
Further, the change of $\mathcal{L}$ due to \prettyref{ln:ADMMProb1} can also be bounded as follows:
\begin{lemma}
\label{lem:ADMM2}
If $\nabla\mathcal{O}$ is $L_\mathcal{O}$-Lipschitz continuous, then \prettyref{ln:ADMMProb1} of \prettyref{alg:ADMM1} will monotonically increase $\mathcal{L}$ by at most:
\begin{align*}
\left[\frac{L_\mathcal{O}+\varrho}{2}-\beta\right]\|\bar{X}_i^{k+1}-\bar{X}_i^k\|^2.
\end{align*}
\end{lemma}
\begin{IEEEproof}
There are two terms related to $\bar{X}_i$, namely $\mathcal{O}(\bar{X}_i)$ and $\varrho/2\|\bar{X}_i(\theta^{k+1})-\bar{X}_i-\lambda_i^k/\varrho\|^2$. Using \prettyref{eq:ADMM1}, the change to the first term can be bounded as:
\small
\begin{align*}
&\mathcal{O}(\bar{X}_i^{k+1})-\mathcal{O}(\bar{X}_i^k)\\
\leq&\nabla\mathcal{O}(\bar{X}_i^k)^T(\bar{X}_i^{k+1}-\bar{X}_i^k)+\frac{L_\mathcal{O}}{2}\|\bar{X}_i^{k+1}-\bar{X}_i^k\|^2\\
=&{\lambda_i^{k+1}}^T(\bar{X}_i^{k+1}-\bar{X}_i^k)+\left[\frac{L_\mathcal{O}}{2}-(\beta-\varrho)\right]\|\bar{X}_i^{k+1}-\bar{X}_i^k\|^2.\labelthis{eq:ADMM3}
\end{align*}
\normalsize
The change to the second term can be bounded as:
\footnotesize
\begin{align*}
&\frac{\varrho}{2}\|\bar{X}_i(\theta^{k+1})-\bar{X}_i^{k+1}-\frac{\lambda_i^k}{\varrho}\|^2-
\frac{\varrho}{2}\|\bar{X}_i(\theta^{k+1})-\bar{X}_i^k-\frac{\lambda_i^k}{\varrho}\|^2\\
=&\frac{\varrho}{2}(\bar{X}_i^k-\bar{X}_i^{k+1})^T
\left[\frac{-2\lambda_i^{k+1}}{\varrho}+\bar{X}_i^{k+1}-\bar{X}_i^k\right]\\
=&-\frac{\varrho}{2}\|\bar{X}_i^{k+1}-\bar{X}_i^k\|^2-{\lambda_i^{k+1}}^T(\bar{X}_i^{k+1}-\bar{X}_i^k).\labelthis{eq:ADMM4}
\end{align*}
\normalsize
We can derive the result to be proved by summing up \prettyref{eq:ADMM3} and \prettyref{eq:ADMM4}.
\end{IEEEproof}
We can then establish the Lyapunov candidate $\mathcal{L}_\kappa^k\triangleq\mathcal{L}(\theta^{k+1},\bar{X}_i,\lambda_i^k,n_i^k,d_i^k,n_{ij}^k,d_{ij}^k)+\kappa\|\bar{X}_i^{k+1}-\bar{X}_i^k\|^2$ and prove its monotonic property below:
\begin{lemma}
\label{lem:ADMM3}
If $\nabla\mathcal{O}$ is $L_\mathcal{O}$-Lipschitz continuous, then the sequence $\{\mathcal{L}_\kappa^k\}$ is monotonically decreasing when:
\begin{align*}
\varrho=\beta\quad\kappa=\beta/4\quad\beta>2\sqrt{2}L_\mathcal{O}.
\end{align*}
\end{lemma}
\begin{IEEEproof}
The other parts of \prettyref{alg:ADMM1} are monotonically decreasing $\mathcal{L}$ except for \prettyref{ln:ADMMProb1} and \prettyref{ln:ADMMProb2}. Combining the results of \prettyref{lem:ADMM1} and \prettyref{lem:ADMM2}, we have:
\footnotesize
\begin{align*}
&\mathcal{L}_\kappa^{k+1}-\mathcal{L}_\kappa^k\leq
\kappa\sum_i\|\bar{X}_i^{k+1}-\bar{X}_i^k\|^2-\kappa\sum_i\|\bar{X}_i^k-\bar{X}_i^{k-1}\|^2+\\
&\frac{2(\beta-\varrho)^2}{\varrho}\sum_i\|\bar{X}_i^{k+1}-\bar{X}_i^k\|^2+
\frac{2(\beta-\varrho+L_\mathcal{O})^2}{\varrho}\sum_i\|\bar{X}_i^k-\bar{X}_i^{k-1}\|^2+\\
&\left[\frac{L_\mathcal{O}+\varrho}{2}-\beta\right]\sum_i\|\bar{X}_i^{k+1}-\bar{X}_i^k\|^2\\
=&\left[\frac{2(\beta-\varrho)^2}{\varrho}+\frac{L_\mathcal{O}+\varrho}{2}-\beta+\kappa\right]\sum_i\|\bar{X}_i^{k+1}-\bar{X}_i^k\|^2+\\
&\left[\frac{2(\beta-\varrho+L_\mathcal{O})^2}{\varrho}-\kappa\right]\sum_i\|\bar{X}_i^k-\bar{X}_i^{k-1}\|^2.
\end{align*}
\normalsize
It can be verified that both terms in the last equation are negative using the parameter choices give above.
\end{IEEEproof}
Next, we show that the sequence $\{\mathcal{L}_\kappa^k\}$ is convergent:
\begin{lemma}
\label{lem:ADMM4}
We denote the following remainder as $\mathcal{R}$:
\begin{align*}
&\mathcal{R}(\theta,n_i,d_i,n_{ij},d_{ij})\triangleq \\
&\gamma\sum_{i}\left[\sum_{x\in X_i}\log(n_ix(\theta)+d_i)+\sum_{z\in Z}\log(-n_iz-d_i)\right]-    \\
&\gamma\sum_{ij}\left[\sum_{x\in X_i}\log(n_{ij}x(\theta)+d_{ij})+\sum_{x\in X_j}\log(-n_{ij}x(\theta)-d_{ij})\right].
\end{align*}
If $\nabla\mathcal{O}$ is $L_\mathcal{O}$-Lipschitz continuous, $\mathcal{R}\geq\underline{\mathcal{R}}$ and $\mathcal{O}\geq\underline{\mathcal{O}}$ are lower-bounded and parameters are chosen according to \prettyref{lem:ADMM3}, then the Lyapunov candidate $\mathcal{L}_\kappa$ is lower-bounded and the sequence $\{\mathcal{L}_\kappa^k\}$ is convergent.
\end{lemma}
\begin{IEEEproof}
We have the following result from \prettyref{eq:ADMM2}:
\small
\begin{align*}
&\mathcal{L}_\kappa^k\geq\underline{\mathcal{R}}+\sum_i\mathcal{O}(\bar{X}_i^k)+\\
&\sum_i\frac{\varrho}{2}\|X_i(\theta^k)-\bar{X}_i^k\|^2+{\lambda_i^k}^T(X_i(\theta^k)-\bar{X}_i^k)\\
=&\underline{\mathcal{R}}+\sum_i\mathcal{O}(\bar{X}_i^k)+\nabla\mathcal{O}(\bar{X}_i^k)^T(X_i(\theta^k)-\bar{X}_i^k)+\\
&\sum_i\frac{\varrho}{2}\|X_i(\theta^k)-\bar{X}_i^k\|^2+
(\beta-\varrho)(\bar{X}_i^{k+1}-\bar{X}_i^k)^T(X_i(\theta^k)-\bar{X}_i^k)\\
\geq&\underline{\mathcal{R}}+\sum_i\mathcal{O}(X_i(\theta^k))-\frac{L_\mathcal{O}}{2}\|X_i(\theta^k)-\bar{X}_i^k\|^2+\\
&\sum_i\frac{\varrho}{2}\|X_i(\theta^k)-\bar{X}_i^k\|^2+
(\beta-\varrho)(\bar{X}_i^{k+1}-\bar{X}_i^k)^T(X_i(\theta^k)-\bar{X}_i^k)\\
\geq&\underline{\mathcal{R}}+\sum_i\underline{\mathcal{O}}+
\frac{\varrho-L_\mathcal{O}}{2}\|X_i(\theta^k)-\bar{X}_i^k\|^2\geq\underline{\mathcal{R}}+\sum_i\underline{\mathcal{O}},
\end{align*}
\normalsize
so the sequence $\{\mathcal{L}_\kappa^k\}$ is monotonically decreasing, lower-bounded, and thus convergent. 
\end{IEEEproof}
\begin{remark}
If time optimality is considered, then \prettyref{lem:ADMM3} holds with the following choice of remainder:
\begin{align*}
&\mathcal{R}(\theta,\Delta t,n_i,d_i,n_{ij},d_{ij})\triangleq w\Delta t-    \\
&\gamma\sum_{i}\log(v_\text{max}\Delta t-\|V_i\|)-\gamma\sum_{i}\log(a_\text{max}\Delta t^2-\|A_i\|)-    \\
&\gamma\sum_{i}\left[\sum_{x\in X_i}\log(n_ix(\theta)+d_i)+\sum_{z\in Z}\log(-n_iz-d_i)\right]-    \\
&\gamma\sum_{ij}\left[\sum_{x\in X_i}\log(n_{ij}x(\theta)+d_{ij})+\sum_{x\in X_j}\log(-n_{ij}x(\theta)-d_{ij})\right].
\end{align*}
\end{remark}
Next, we establish the convergence guarantee for a convergent sub-sequence. The first-order critical point should satisfy the following conditions:
\begin{align*}
&\left\|\FPP{\mathcal{L}}{\lambda_i}\right\|=\|X_i(\theta)-\bar{X}_i\|=0\labelthis{eq:ADMM1KKT1}\\
&\left\|\FPP{\mathcal{L}}{\bar{X}_i}\right\|=\|\nabla\mathcal{O}(\bar{X}_i)-\lambda_i\|=0\labelthis{eq:ADMM1KKT2}\\
&\left\|\nabla\mathcal{L}\circ\exp\FIVE{\Theta}{\exp_{n_i}(0)}{d_i}{\exp_{n_{ij}}(0)}{d_{ij}}\right\|=0,\labelthis{eq:ADMM1KKT3}
\end{align*}
where we define $\mathcal{L}\circ\exp$ with each $n_i$ (w.r.t. $n_{ij}$) replaced by $\exp_{n_i}(v_i)$ (w.r.t. $\exp_{n_{ij}}(v_{ij})$).
\begin{lemma}
\label{lem:ADMM5}
If $\nabla\mathcal{O}$ is $L_\mathcal{O}$-Lipschitz continuous, both $\mathcal{R}$ and $\mathcal{O}$ are lower-bounded and parameters are chosen according to \prettyref{lem:ADMM3}, then every accumulation point of the sequence $\{\FOUR{\Theta^k}{\bar{X}_i^k}{P_i^k}{P_{ij}^k}\}$ generated by \prettyref{alg:ADMM1} is a critical point. Suppose the sequence is convergent, the local convergence speed of $\|\nabla\mathcal{L}\circ\exp\|$ is $\mathcal{O}(1/\sqrt{K})$.
\end{lemma}
\begin{IEEEproof}
The sequence $\{\mathcal{L}_\kappa\FIVE{\Theta^k}{\bar{X}_i^k}{\lambda_i^k}{P_i^k}{P_{ij}^k}\}$ is convergent due to \prettyref{lem:ADMM4}. We assume the accumulation point is $\FOUR{\bar{\Theta}}{\bar{\bar{X}}_i}{\bar{P}_i}{\bar{P}_{ij}}$ with convergent sub-sequence identified by the index subset: $\mathcal{K}\subset\{k|k=1,2,\cdots\}$.

\TE{Applying \prettyref{lem:AM1} and \prettyref{lem:ADMM3}:} The reduction of $\mathcal{L}_\kappa$ over one iteration of \prettyref{alg:ADMM1} can be bounded as:
\begin{align*}
&\mathcal{L}_\kappa^{k+1}+c\|\Theta^{k+1}-\Theta^k\|^2+\\
&c\sum_i\|v_i^k\|^2+\|d_i^{k+1}-d_i^k\|^2+c\sum_{ij}\|v_{ij}^k\|^2+\|d_{ij}^{k+1}-d_{ij}^k\|^2-\\
&\left[\frac{2(\beta-\varrho)^2}{\varrho}+\frac{L_\mathcal{O}+\varrho}{2}-\beta+\kappa\right]\sum_i\|\bar{X}_i^{k+1}-\bar{X}_i^k\|^2-\\
&\left[\frac{2(\beta-\varrho+L_\mathcal{O})^2}{\varrho}-\kappa\right]\sum_i\|\bar{X}_i^k-\bar{X}_i^{k-1}\|^2\leq\mathcal{L}_\kappa^k,\labelthis{eq:ADMM5}
\end{align*}
from which we conclude that $\lim_{k\to\infty}\|\Theta^{k+1}-\Theta^k\|=0$, $\lim_{k\to\infty}\|v^k\|=0$, and $\lim_{k\to\infty}\|d^{k+1}-d^k\|=0$ for any $P$, and $\lim_{k\to\infty}\|\bar{X}_i^{k+1}-\bar{X}_i^k\|=0$. \prettyref{eq:ADMM1KKT3} is satisfied at the accumulation point following the same reasoning as \prettyref{lem:AM2}. Taking limits on both sides of \prettyref{eq:ADMM2} and we have:
\begin{align*}
&\lim_{k\to\infty}\|\lambda^{k+1}-\nabla\mathcal{O}(\bar{X}_i^{k+1})\|\\
\leq&\lim_{k\to\infty}\|\lambda^{k+1}-\nabla\mathcal{O}(\bar{X}_i^k)\|+\|\nabla\mathcal{O}(\bar{X}_i^{k+1})-\nabla\mathcal{O}(\bar{X}_i^k)\|\\
\leq&L_\mathcal{O}\lim_{k\to\infty}\|\bar{X}_i^{k+1}-\bar{X}_i^k\|=0,
\end{align*}
from which \prettyref{eq:ADMM1KKT2} follows. We then take limits on both sides of \prettyref{ln:ADMMProb2} and we have $\lim_{k\to\infty}\|X_i(\theta^k)-\bar{X}_i^k\|=0$, from which \prettyref{eq:ADMM1KKT1} follows.

\TE{Convergence Speed:} Summing up \prettyref{eq:ADMM5} over $K$ iterations and we have:
\small
\begin{align*}
&K\min_{k=1,\cdots,K}\left[\|\bar{X}_i^{k+1}-\bar{X}_i^k\|^2+\|\bar{X}_i^k-\bar{X}_i^{k-1}\|^2\right]\\
\leq&\frac{\mathcal{L}_\kappa^1-\underline{\mathcal{L}}_\kappa}{\min(-\left[\frac{2(\beta-\varrho)^2}{\varrho}+\frac{L_\mathcal{O}+\varrho}{2}-\beta+\kappa\right],-\left[\frac{2(\beta-\varrho+L_\mathcal{O})^2}{\varrho}-\kappa\right])}.
\end{align*}
\normalsize
We plug these results into \prettyref{eq:ADMM1} and \prettyref{lem:ADMM1} to derive:
\footnotesize
\begin{align*}
&\|X_i(\theta^{k+1})-\bar{X}_i^{k+1}\|^2=\frac{1}{\varrho^2}\|\lambda_i^{k+1}-\lambda_i^k\|^2\\
\leq&\frac{2(\beta-\varrho)^2}{\varrho^2}\|\bar{X}_i^{k+1}-\bar{X}_i^k\|^2+\frac{2(\beta-\varrho+L_\mathcal{O})^2}{\varrho^2}\|\bar{X}_i^k-\bar{X}_i^{k-1}\|^2\\
\leq&\frac{\max(2(\beta-\varrho)^2,2(\beta-\varrho+L_\mathcal{O})^2)}{\varrho^2}\left[\|\bar{X}_i^{k+1}-\bar{X}_i^k\|^2+\|\bar{X}_i^k-\bar{X}_i^{k-1}\|^2\right]\\
&\|\lambda_i^k-\nabla\mathcal{O}(\bar{X}_i^k)\|^2\leq2\|\lambda_i^{k+1}-\nabla\mathcal{O}(\bar{X}_i^k)\|^2+2\|\lambda_i^{k+1}-\lambda_i^k\|^2\\
\leq&4(\beta-\varrho)^2\|\bar{X}_i^{k+1}-\bar{X}_i^k\|^2+2(\beta-\varrho+L_\mathcal{O})^2\|\bar{X}_i^k-\bar{X}_i^{k-1}\|^2\\
\leq&\max(4(\beta-\varrho)^2,2(\beta-\varrho+L_\mathcal{O})^2)\left[\|\bar{X}_i^{k+1}-\bar{X}_i^k\|^2+\|\bar{X}_i^k-\bar{X}_i^{k-1}\|^2\right],
\end{align*}
\normalsize
which establishes the $\mathcal{O}(1/\sqrt{K})$ convergence speed of \prettyref{eq:ADMM1KKT1} and \prettyref{eq:ADMM1KKT2}. The local convergence speed of \prettyref{eq:ADMM1KKT3} follows the same reasoning as \prettyref{lem:AM2}.
\end{IEEEproof}
Finally, we establish whole sequence convergence and thus local convergence speed assuming K{\L} property of $\mathcal{L}\circ\exp$.
\begin{lemma}
\label{lem:ADMM6}
If $\nabla\mathcal{O}$ is $L_\mathcal{O}$-Lipschitz continuous, $\mathcal{R}$ and $\mathcal{O}$ are lower-bounded and pertains the K{\L} property (\prettyref{def:KL}), and the sequence $\{\FOUR{\Theta^k}{\bar{X}_i^k}{P_i^k}{P_{ij}^k}\}$ generated by \prettyref{alg:ADMM1} has a finite accumulation point, then the sequence is convergent.
\end{lemma}
\begin{IEEEproof}
We denote $\FIVE{\bar{\Theta}}{\bar{\bar{X}}_i}{\bar{\lambda}_i}{\bar{P}_i}{\bar{P}_{ij}}$ as the accumulation point. By invoking \prettyref{lem:ADMM5}, we know that $\nabla\mathcal{L}\circ\exp=0$. W.l.o.g., we assume $\mathcal{L}\FIVE{\bar{\Theta}}{\bar{\bar{X}}_i}{\bar{\lambda}_i}{\bar{P}_i}{\bar{P}_{ij}}=0$. By continuity, we can choose a small neighborhood $B_{2\rho}\subset\DOM\mathcal{L}$ where $\|\nabla\mathcal{L}\circ\exp\|<\epsilon$ and the the K{\L} property holds. We further denote $\nabla\mathcal{L}\circ\exp$ as $L$-Lipschitz continuous and every $X_i$ is $L_{X_i}$-Lipschitz continuous within $B_{2\rho}$. Finally, in a sufficiently small local neighborhood, we can setup strong equivalence between metrics in tangent space and ambient space for any point in $B_{2\rho}$. If we start from $\FIVE{\Theta^k}{\bar{X}_i^k}{\lambda_i^k}{P_i^k}{P_{ij}^k}\in B_\rho$, we choose sufficiently large $k_0$ such that the subsequent point $\FIVE{\Theta^{k+1}}{\bar{X}_i^{k+1}}{\lambda_i^{k+1}}{P_i^{k+1}}{P_{ij}^{k+1}}\in B_{2\rho}$ for every $k\geq k_0$ due to \prettyref{lem:AM2}. W.l.o.g., we can assume $k_0=0$. Since there is a sub-sequence converging to $\FIVE{\bar{\Theta}}{\bar{\bar{X}}_i}{\bar{\lambda}_i}{\bar{P}_i}{\bar{P}_{ij}}$, we can choose large enough $k\geq k_0$ (W.l.o.g., we can assume $k_0=0$) so that:
\footnotesize
\begin{align}
\FIVE{\Theta^0}{\bar{X}_i^0}{\lambda_i^0}{P_i^0}{P_{ij}^0}\in B_{\rho-L_1/L_2\phi(\mathcal{L}\FIVE{\Theta^0}{\bar{X}_i^0}{\lambda_i^0}{P_i^0}{P_{ij}^0}}.
\end{align}
Next, we show that every subsequent point $\FIVE{\Theta^k}{\bar{X}_i^k}{\lambda_i^k}{P_i^k}{P_{ij}^k}\in B_\rho$ lies in by induction. If we already have this property for all previous points, i.e., $\FIVE{\Theta^0}{\bar{X}_i^0}{\lambda_i^0}{P_i^0}{P_{ij}^0},\cdots,\FIVE{\Theta^k}{\bar{X}_i^k}{\lambda_i^k}{P_i^k}{P_{ij}^k}\in B_\rho$, then the following holds for $\FIVE{\Theta^{k+1}}{\bar{X}_i^{k+1}}{\lambda_i^{k+1}}{P_i^{k+1}}{P_{ij}^{k+1}}\in B_{\rho}$:
\scriptsize
\begin{align*}
&\frac{\mathcal{L}\FIVE{\Theta^k}{\bar{X}_i^k}{\lambda_i^k}{P_i^k}{P_{ij}^k}-\mathcal{L}\FIVE{\Theta^{k+1}}{\bar{X}_i^{k+1}}{\lambda_i^{k+1}}{P_i^{k+1}}{P_{ij}^{k+1}}}
{\phi(\mathcal{L}\FIVE{\Theta^k}{\bar{X}_i^k}{\lambda_i^k}{P_i^k}{P_{ij}^k})-\phi(\mathcal{L}\FIVE{\Theta^{k+1}}{\bar{X}_i^{k+1}}{\lambda_i^{k+1}}{P_i^{k+1}}{P_{ij}^{k+1}})}\\
\leq&1/\phi'(\mathcal{L}\FIVE{\Theta^k}{\bar{X}_i^k}{\lambda_i^k}{P_i^k}{P_{ij}^k})\\
\leq&\|\nabla\mathcal{L}(\Theta^k,\bar{X}_i^k,\lambda_i^k,\exp_{n_i^k}(0),d_i^k,\exp_{n_{ij}^k}(0),d_{ij}^k)\|\\
=&\Delta+\sum_i\|\varrho(\bar{X}_i^k-X_i(\theta^k))+\nabla\mathcal{O}(\bar{X}_i^k)-\lambda_i^k\|+\sum_i\|X_i(\theta^k)-\bar{X}_i^k\|\\
\leq&\Delta+\sum_i\|\nabla\mathcal{O}(\bar{X}_i^k)-\lambda_i^{k+1}+\lambda_i^{k+1}-\lambda^k\|+(\varrho+1)\|X_i(\theta^k)-\bar{X}_i^k\|\\
\leq&\Delta+\sum_i\|\lambda_i^{k+1}-\lambda^k\|+(\beta-\varrho)\|\bar{X}_i^{k+1}-\bar{X}_i^k\|+(\varrho+1)\|X_i(\theta^k)-\bar{X}_i^k\|\\
\leq&\Delta+\sum_i\|\lambda_i^{k+1}-\lambda^k\|+(\beta-\varrho)\|\bar{X}_i^{k+1}-\bar{X}_i^k\|+\\
&(\varrho+1)\left[\|X_i(\theta^{k+1})-\bar{X}_i^{k+1}\|+\|X_i(\theta^{k+1})-X_i(\Theta^k)\|+\|\bar{X}_i^{k+1}-\bar{X}_i^k\|\right]\\
\leq&\Delta+\sum_i\frac{2\varrho+1}{\varrho}\|\lambda_i^{k+1}-\lambda^k\|+\\
&(\beta+1)\|\bar{X}_i^{k+1}-\bar{X}_i^k\|+(\varrho+1)L_{X_i}\|\Theta_i^{k+1}-\Theta^k\|\\
\leq&L_1\|\FIVE{\Theta^{k+1}}{\bar{X}_i^{k+1}}{\lambda_i^{k+1}}{P_i^{k+1}}{P_{ij}^{k+1}}-\FIVE{\Theta^k}{\bar{X}_i^k}{\lambda_i^k}{P_i^k}{P_{ij}^k}\|,
\end{align*}
\normalsize
where we define:
\footnotesize
\begin{align*}
\Delta\triangleq&\left[\frac{L}{2\gamma(1-c)}+\sum_iL+\sum_{ij}L\right]\|\Theta^{k+1}-\Theta^k\|+\\
&\frac{L\bar{L}}{2\gamma(1-c)}\left[\sum_i\|P_i^{k+1}-P_i^k\|+\sum_{ij}\|P_{ij}^{k+1}-P_{ij}^k\|\right]\\
L_1=&\max\TWOC{\frac{L}{2\gamma(1-c)}+\sum_iL+\sum_{ij}L+(\varrho+1)\sum_iL_{X_i}}
{\max\THREE{\frac{L\bar{L}}{2\gamma(1-c)}}{\beta+1}{\frac{2\varrho+1}{\varrho}}}.
\end{align*}
\normalsize
We have used \prettyref{eq:ADMM1} in the forth inequality and the $L_{X_i}$-Lipschitz smoothness in the sixth inequality. We further have the following estimate of function decrease due to \prettyref{lem:ADMM5}:
\footnotesize
\begin{align*}
&\mathcal{L}\THREE{\Theta^k}{P_i^k}{P_{ij}^k}-\mathcal{L}\THREE{\Theta^{k+1}}{P_i^{k+1}}{P_{ij}^{k+1}}\\
\geq&c\|\Theta^{k+1}-\Theta^k\|^2+c\underline{L}^2\sum_i\|n_i^{k+1}-n_i^k\|^2+c\sum_i\|d_i^{k+1}-d_i^k\|^2+\\
&c\underline{L}^2\sum_{ij}\|n_{ij}^{k+1}-n_{ij}^k\|^2+c\sum_{ij}\|d_{ij}^{k+1}-d_{ij}^k\|^2-\\
&\left[\frac{2(\beta-\varrho)^2}{\varrho}+\frac{L_\mathcal{O}+\varrho}{2}-\beta+\kappa\right]\|\bar{X}_i^{k+1}-\bar{X}_i^k\|^2-\\
&\left[\frac{2(\beta-\varrho+L_\mathcal{O})^2}{\varrho}-\kappa\right]\|\bar{X}_i^k-\bar{X}_i^{k-1}\|^2\\
\geq&L_2\|\FIVE{\Theta^{k+1}}{\bar{X}_i^{k+1}}{\lambda_i^{k+1}}{P_i^{k+1}}{P_{ij}^{k+1}}-\FIVE{\Theta^k}{\bar{X}_i^k}{\lambda_i^k}{P_i^k}{P_{ij}^k}\|^2,
\end{align*}
\normalsize
where we define:
\small
\begin{align*}
L_2=\min\THREEC{\min(c,c{\underline{L}}^2)}{-\left[\frac{2(\beta-\varrho)^2}{\varrho}+\frac{L_\mathcal{O}+\varrho}{2}-\beta+\kappa\right]}{-\left[\frac{2(\beta-\varrho+L_\mathcal{O})^2}{\varrho}-\kappa\right]}.
\end{align*}
\normalsize
The remaining argument is identical to \prettyref{lem:AM3}.
\end{IEEEproof}